\documentclass[11pt]{article}

\usepackage[preprint]{acl}

\usepackage{times}
\usepackage{latexsym}

\usepackage[T1]{fontenc}

\usepackage[utf8]{inputenc}

\usepackage{microtype}

\usepackage{inconsolata}

\usepackage{graphicx}
\usepackage{amsmath}
\usepackage{booktabs}
\usepackage{colortbl}
\usepackage[table]{xcolor}
\definecolor{mathcolor}{rgb}{0.85, 0.91, 0.98}
\usepackage{enumitem}
\usepackage{multirow}
\usepackage{bbm}
\usepackage{tabularx}
\newcolumntype{Y}{>{\centering\arraybackslash}X}
\usepackage{subcaption}
\usepackage{multicol}
\usepackage{enumitem}
\usepackage{titlesec}
\titlespacing*{\section}      {0pt}{1.2ex plus .2ex minus .2ex}{0.8ex plus .1ex}
\titlespacing*{\subsection}   {0pt}{1.0ex plus .2ex minus .2ex}{0.6ex plus .1ex}
\titlespacing*{\subsubsection}{0pt}{0.8ex plus .1ex minus .1ex}{0.5ex plus .1ex}
\title{From Meta-Thought to Execution: Cognitively Aligned Post-Training for Generalizable and Reliable LLM Reasoning}

\author{
  \textbf{Shaojie Wang\textsuperscript{1}} \quad
  \textbf{Liang Zhang\textsuperscript{1}}\thanks{\ Corresponding author.} \\
  \\
  \texttt{\{shaojiewang, liangzhang\}@hkustgz.edu.cn} \\
  \\
  \textsuperscript{1} Hong Kong University of Science and Technology (Guangzhou) \\
}

\begin{document}
\maketitle
\begin{abstract}
Current LLM post-training methods optimize complete reasoning trajectories through Supervised Fine-Tuning (SFT) followed by outcome-based Reinforcement Learning (RL). While effective, a closer examination reveals a fundamental gap: this approach does not align with how humans actually solve problems. Human cognition naturally decomposes problem-solving into two distinct stages: first acquiring abstract strategies (\textit{i.e.}, meta-knowledge) that generalize across problems, then adapting them to specific instances. In contrast, by treating complete trajectories as basic units, current methods are inherently problem-centric, entangling abstract strategies with problem-specific execution. To address this misalignment, we propose a cognitively-inspired framework that explicitly mirrors the two-stage human cognitive process. Specifically, \textbf{C}hain-\textbf{o}f-\textbf{M}eta-\textbf{T}hought (\textbf{CoMT}) focuses supervised learning on abstract reasoning patterns without specific executions, enabling acquisition of generalizable strategies. \textbf{C}onfidence-\textbf{C}alibrated \textbf{R}einforcement \textbf{L}earning (\textbf{CCRL}) then optimizes task adaptation via confidence-aware rewards on intermediate steps, preventing overconfident errors from cascading and improving execution reliability. Experiments across four models and ten benchmarks show 2.10\% and 3.86\% improvements in-distribution and out-of-distribution respectively over standard methods, while remaining highly robust to variations in teacher model selection, optimization methods, and symbolic perturbations.
\end{abstract}


\section{Introduction}
\label{sec:intro}
Mathematical reasoning, as a crucial cognitive skill that supports problem-solving across numerous scientific and real-world applications, has attracted significant attention in recent years~\citep{llm4mathreasoning,llm_mathematical_survey}.
To this end, post-training is one of the most critical techniques for enhancing this capability in Large Language Models (LLMs). The dominant paradigm typically follows a two-stage pipeline: Chain of Thought Supervised Fine-Tuning (CoT-SFT) teaches models to generate step-by-step reasoning trajectories, followed by Reinforcement Learning (RL) that optimizes responses based on outcome feedback, typically final answer correctness~\citep{sft+rl_training,llm_comparative_study,llm_post-training_dive_survey}. This CoT-SFT+RL framework has proven highly effective for improving mathematical reasoning performance~\citep{plan_tuning,srl}.


Despite these advances, an important question remains unexplored: how does this popular training paradigm align with the way humans actually solve problems? This alignment matters because human problem-solving exhibits precisely the property we seek in LLMs: robust generalization from limited experience to novel situations~\citep{principles_of_generalization}. Understanding the correspondence between LLM training approaches and human problem-solving processes can motivate us to rethink existing post-training paradigms and inspire the development of more effective methodologies.

Cognitive science reveals the key to human problem-solving effectiveness: abstraction---the ability to extract general principles from previous experiences and apply them to novel situations~\citep{abstraction_in_ai}. When encountering a new problem, humans do not start from scratch; instead, they recognize its underlying structure, retrieve relevant strategies, and adapt them to the specific context at hand~\citep{structure_mapping,disentangling_abstract}. This cognitive process can be naturally decomposed into two stages: first, acquiring abstract problem-solving knowledge, what we term \textbf{meta-knowledge}, that generalizes across 
different problems; second, adapting this meta-knowledge to execute solutions for specific instances~\citep{schema_induction,transfer_of_cognitive_skill,causal_transfer,structural_knowledge_transfer}. 

Comparing the CoT-SFT+RL paradigm to this human cognitive process reveals misalignments in how models acquire and apply problem-solving knowledge. Specifically, current methods are inherently 
problem-centric: CoT-SFT encourages imitation of complete reasoning trajectories for individual problems, while RL optimizes based on answer correctness of specific instances. The learning signal primarily reflects what works for individual problems rather than explicitly supervising the acquisition of meta-knowledge that generalizes across problem classes. Moreover, CoT-SFT+RL pipeline operates on complete chain-of-thought (CoT) sequences as the basic unit of supervision, inherently entangling abstract strategy formation with problem-specific execution into a single undifferentiated signal. This conflation makes it difficult for models to separately internalize what general strategies to acquire versus how to adapt and apply them to specific instances, two distinct processes that cognitive research shows to characterize human skill acquisition~\citep{sweller1988cognitive}. These misalignments highlight a fundamental gap between how current methods train LLM models and how humans acquire transferable problem-solving knowledge.

To bridge this gap, we propose a new post-training paradigm that explicitly decomposes the learning process into two distinct stages, mirroring the human cognitive process. Our approach reconceptualizes post-training: rather than optimizing for problem-level performance, we explicitly separate meta-knowledge acquisition from task-specific adaptation. First, \textbf{meta-knowledge acquisition} focuses on learning abstract, task-agnostic strategies, such as general solution strategies applicable to problem classes sharing common patterns without entangling problem-specific execution details. Second, \textbf{task adaptation} focuses on applying the acquired 
meta-knowledge to solve specific problems through problem-optimized reasoning steps. By decoupling these two processes, our framework addresses limitations of the CoT-SFT+RL paradigm: it shifts from problem-centric optimization to strategy-centric learning, and disentangles the supervision of abstract reasoning strategies from concrete problem-specific execution.


To operationalize this framework, we introduce two complementary techniques that correspond to the two stages of human cognition. For \textbf{meta-knowledge acquisition}, we propose \textbf{C}hain-\textbf{o}f-\textbf{M}eta-\textbf{T}hought (\textbf{CoMT}) Supervised Fine-Tuning, which focuses on learning abstract reasoning patterns rather than imitating complete chain-of-thought solutions with specific execution. Crucially, CoMT supervision explicitly replaces all numerical values with variable names, excluding problem-specific execution details and encouraging the model to internalize abstract reasoning knowledge that is distinct from mere length reduction or stylistic regularization of CoT. Subsequently, for \textbf{task adaptation}, we face a distinct challenge: while CoMT helps models acquire abstract strategies, applying these strategies to specific problems requires reliable multi-step execution. However, execution reliability is undermined when models are overconfident in incorrect intermediate computations—such errors cascade through subsequent 
steps, compounding into wrong final answers~\citep{uesato2022solving}. 
To address this, we employ \textbf{C}onfidence-\textbf{C}alibrated \textbf{R}einforcement \textbf{L}earning (\textbf{CCRL}) to optimize models' confidence calibration during execution. Unlike outcome-based RL, CCRL incorporates confidence-aware rewards on intermediate computations: it rewards high-confidence correct predictions while penalizing high-confidence incorrect predictions. This calibration mechanism encourages models to be confident when correct and uncertain when erring, thus preventing overconfident errors from compounding across reasoning steps and enhancing execution reliability.

Together, this framework instantiates the core principles of human cognition within standard LLM post-training pipelines: abstraction and adaptation. The design maintains compatibility with existing post-training infrastructure while shifting the learning objective from problem-level imitation to strategy-level generalization. Our contributions are as follows:
\begin{itemize}[itemsep=2pt, topsep=2pt, parsep=0pt, partopsep=0pt, leftmargin=*]
    \item We propose a cognitively-inspired post-training framework that decomposes reasoning into meta-knowledge acquisition and task adaptation, shifting from problem-centric optimization to strategy-centric learning aligned with human cognition.
    \item We introduce Chain-of-Meta-Thought (CoMT) supervised fine-tuning that focuses on learning abstract problem-solving strategies without problem-specific execution details, and Confidence-Calibrated Reinforcement Learning (CCRL) that incorporates confidence-aware rewards to encourage confidence calibration—being confident when correct and uncertain when making errors, thereby promoting reliable strategy execution.
    \item Extensive experiments demonstrate that our framework achieves superior performance and generalization compared to baselines, while requiring much fewer computational resources and maintaining compatibility with post-training infrastructure.
\end{itemize}

\section{Related Work}
\subsection{Post-training for LLM Reasoning}
Post-training has become a key lever for improving LLM reasoning beyond pre-training~\citep{llm_post-training_dive_survey,llm_comparative_study}. The dominant paradigm is a two-stage pipeline: CoT-SFT teaches step-by-step reasoning trajectories~\citep{cot}, followed by RL that optimizes responses with outcome-based rewards~\citep{sft+rl_training}. Variants refine this framework along several axes—self-bootstrapping~\citep{bosststrapp_reasoning_with_reasoning}, process-level reward modeling~\citep{verify_step_by_step,groundedprm}, uncertainty-based self-feedback~\citep{post_training_self_feedback}, unified SFT–RL training~\citep{uft}, dynamic signal selection~\citep{hpt}, and pure-RL recipes like DeepSeek-R1~\citep{deepseek_r1}. All of these still treat each problem as the unit of optimization. We instead decompose post-training into meta-knowledge acquisition and task adaptation, shifting from problem-centric to strategy-centric learning.

\subsection{Human Cognitive Processes in Problem-Solving}
A hallmark of human intelligence is abstraction: extracting general principles from experience and applying them to new situations~\citep{structure_mapping,abstract_rule}. Cognitive science consistently describes problem-solving as a two-stage process—first acquiring abstract, transferable strategies, then adapting them to specific instances~\citep{schema_induction,transfer_of_cognitive_skill,categorization_representation}. Neuroscience supports this view: the prefrontal cortex implements meta-learning over abstract task structures~\citep{meta_learning_system}, and humans generalize rapidly by recombining compositional primitives~\citep{compositional_generalization,analogical_mapping}. Current LLM post-training, by contrast, optimizes complete reasoning trajectories as monolithic units. Our approach aligns post-training with this decomposed structure of human problem-solving.

\section{Methodology}

\begin{figure}[t]
  \centering
  \includegraphics[width=0.95\columnwidth]{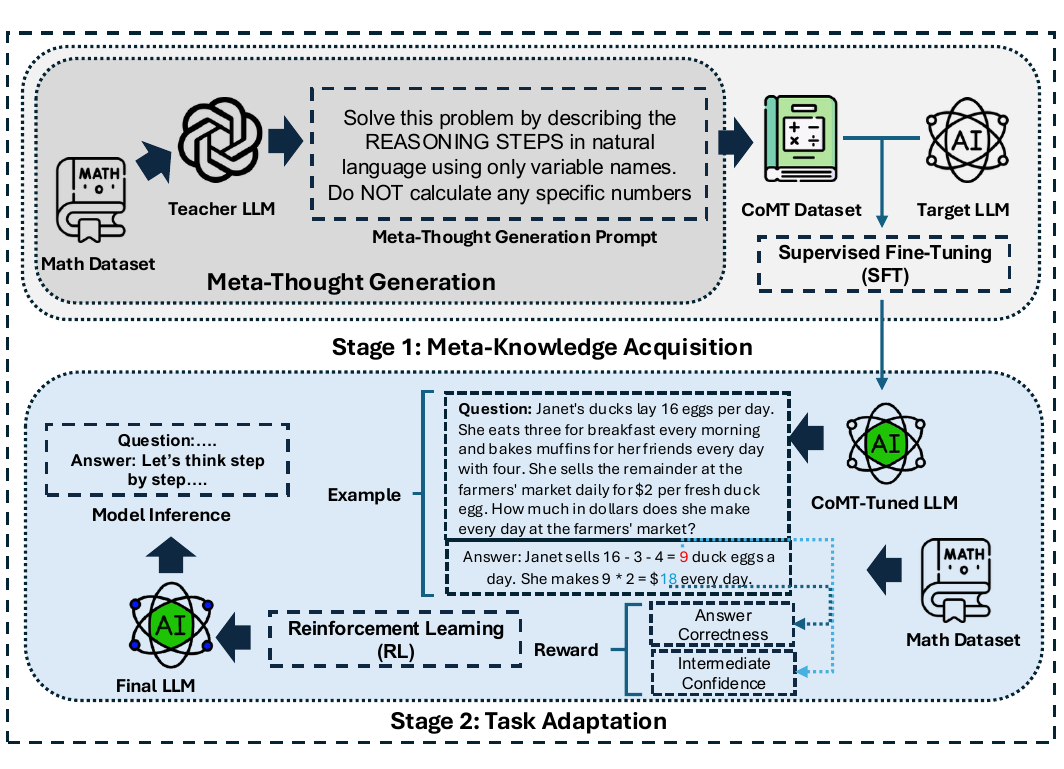}
 \caption{Overview of our two-stage post-training framework with a concrete example from GSM8K. \textbf{Stage 1 (Meta-Knowledge Acquisition):} A teacher LLM generates abstract meta-thoughts excluding numerical calculations, which are used for CoMT supervised fine-tuning. \textbf{Stage 2 (Task Adaptation):} The CoMT-tuned model undergoes RL with rewards combining answer correctness and intermediate confidence scores (highlighted in the example: confidence on $9$ and correctness on $18$).}
  \label{fig:framework}
  \vskip -0.2in
\end{figure}

Our framework decomposes LLM reasoning post-training into two cognitively-inspired stages that mirror human problem-solving: \textbf{meta-knowledge acquisition} and \textbf{task adaptation}. The first stage focuses on learning abstract reasoning patterns without problem-specific execution details. The second stage optimizes how models apply this acquired meta-knowledge to solve specific instances, incorporating confidence-aware feedback to enhance execution reliability through confidence calibration. Figure~\ref{fig:framework} illustrates the overall pipeline.

\subsection{Meta-Knowledge Acquisition}
\label{comt_data_generation}
Chain of Thought Supervised Fine-Tuning (CoT-SFT) teaches models to imitate complete reasoning trajectories, including both abstract problem-solving strategies and concrete execution details. In the mathematical reasoning setting considered in this work, 
this means conflating general solution strategies with specific numerical computations. As discussed in Section~\ref{sec:intro}, this conflation makes it difficult for models to extract and internalize the reusable meta-knowledge that generalizes across different problems. Inspired by how humans acquire problem-solving strategies—understanding general approaches rather than memorizing specific solutions—we propose Chain-of-Meta-Thought (CoMT) supervised fine-tuning.

We first generate meta-thought trajectories using a teacher model, then train the target model to produce similar abstract reasoning patterns. The data generation process is designed to elicit abstract reasoning: specifically, the constructed CoMT training data emphasizes describing abstract \textit{reasoning steps} rather than solving problems.

\textbf{Data Generation with Teacher Model.} 
For each problem $q$ in the training dataset, we query a stronger teacher LLM $\mathcal{M}_{\text{teacher}}$ with the following prompt:

\begin{quote}
\textit{``Solve this problem by describing the REASONING STEPS in natural language using only variable names. Do NOT calculate any specific numbers. Question: [Problem text]''}
\end{quote}

This prompt explicitly instructs the teacher model to focus on the abstract problem-solving strategy rather than numerical computation. The teacher generates a meta-thought trajectory $\tau_{\text{meta}}$ that describes each reasoning step in natural language and uses variable names instead of specific numbers.

Through this process, we construct a dataset $\mathcal{D}_{\text{CoMT}} = \{(q_i, \tau_{\text{meta}}^i)\}_{i=1}^N$ where each problem is paired with its corresponding meta-thought trajectory generated by the teacher model.

\textbf{Training with CoMT Data.}
We then fine-tune the target model parameterized by $\theta$ on $\mathcal{D}_{\text{CoMT}}$ using the same prompting format. During training, the model learns to generate meta-thought trajectories when given the CoMT prompt. 

Let $\tau_{\text{meta}} = (s_1, s_2, \ldots, s_{|\tau_{\text{meta}}|})$ denote the sequence of tokens in a meta-thought trajectory, where $s_t \in \mathcal{V}$ represents the $t$-th token from the vocabulary $\mathcal{V}$. The training objective is standard supervised learning:
\begin{equation}
\mathcal{L}_{\text{CoMT}} = -{E}_{(q, \tau_{\text{meta}}) \sim \mathcal{D}_{\text{CoMT}}} 
\left[ \sum_{t=1}^{|\tau_{\text{meta}}|} \log \pi_\theta(s_t | q, s_{<t}) \right]
\end{equation}
where $\pi_\theta(s_t | q, s_{<t})$ denotes the probability of generating token $s_t$ given the problem $q$ and previous tokens $s_{<t}$. Note that the problem $q$ implicitly includes the CoMT prompting instruction.


\subsection{Task Adaptation}
After acquiring meta-knowledge through CoMT, models must apply these abstract strategies to solve specific problem instances. In mathematical reasoning, this requires reliable multi-step execution where intermediate computations build upon each other. However, a critical challenge undermines the reliability of strategy execution: overconfident errors in intermediate steps cascade through reasoning processes, compounding into incorrect final answers~\citep{uesato2022solving}. While outcome-based RL optimizes for final answer correctness, it rewards any reasoning path that reaches the right conclusion regardless of intermediate step reliability, leaving models without the confidence awareness to recognize when their computations are unreliable. To address this, we propose Confidence-Calibrated Reinforcement Learning (CCRL) that incorporates confidence-aware rewards on intermediate steps, encouraging models to be confident when correct and uncertain when erring, thereby enhancing execution reliability by preventing overconfident errors from cascading.
 
\subsubsection{Identifying Computed Intermediate Results}

During RL training, given a problem $q$, the model generates a reasoning trajectory $y = (y_1, y_2, \ldots, y_T)$, where each $y_t \in \mathcal{V}$ is a token from the vocabulary and $T$ is the sequence length. Unlike the meta-thought trajectories $\tau_{\text{meta}}$ from CoMT training which describe abstract reasoning steps, these trajectories $y$ contain complete solutions with specific numerical computations leading to a final answer.

To measure confidence in intermediate reasoning steps, we focus on the numerical tokens within the generated trajectory, as these intermediate computations are where overconfident errors originate and cascade through subsequent steps. Let $\mathcal{N}_q$ denote the set of numbers appearing in the problem statement $q$. For each numerical token $y_t$ in the trajectory, we classify it into one of two categories: \textbf{Extracted numbers} are numerical tokens that directly appear in the problem, i.e., $y_t \in \mathcal{N}_q$; \textbf{Computed numbers} are numerical tokens produced through intermediate calculations, i.e., $y_t \notin \mathcal{N}_q$.

We focus specifically on computed numbers, as these represent the model's intermediate computational results. Let $\mathcal{C}$ denote the set of indices corresponding to computed number tokens: $\mathcal{C} = \{t : y_t \text{ is numeric and } y_t \notin \mathcal{N}_q\}$.

\subsubsection{Entropy-Based Confidence}

For each computed number token at position $t \in \mathcal{C}$, we measure the model's confidence using the entropy of its predictive distribution at that step. Let $l_t \in {R}^{|\mathcal{V}|}$ denote the model's logits at position $t$. The probability distribution over the vocabulary $\mathcal{V}$ is: 
\begin{equation}
\label{entropy_for_all_tokens}
\pi_\theta(v | q, y_{<t}) = \frac{\exp(l_t[v])}{\sum_{v' \in \mathcal{V}} \exp(l_t[v'])}
\end{equation}

where $l_t[v]$ denotes the logit for token $v$ at position $t$. The entropy at this position is: 
\begin{equation}
\label{entropy_for_one_position}
H_t = -\sum_{v \in \mathcal{V}} \pi_\theta(v | q, y_{<t}) \log \pi_\theta(v | q, y_{<t})
\end{equation}

Lower entropy indicates higher confidence, as the probability mass is concentrated on fewer vocabulary tokens. To obtain a single confidence measure for the entire reasoning trajectory, we use the maximum entropy across all computed numbers: 
\begin{equation}
H_{\max} = \max_{t \in \mathcal{C}} H_t
\end{equation}
This measurement captures the model's uncertainty at its least confident computational step.

\subsubsection{Confidence-Calibrated Reward Function}
\label{sec:ccrl_reward}

Using the entropy measure $H_{\max}$, we define the confidence-calibrated reward:
\begin{equation}
\label{eq:confidence_reward}
r_{\text{confidence}}(q, y) = \begin{cases}
\alpha \cdot e^{-H_{\max}} & \text{if correct} \\
-\beta \cdot e^{-H_{\max}} & \text{if incorrect}
\end{cases}
\end{equation}
where $\alpha > 0$ and $\beta > 0$ control the strength of confidence rewards and penalties. The exponential term $e^{-H_{\max}}$ transforms entropy into a confidence score in $(0, 1]$: lower entropy (higher confidence) yields values closer to 1, resulting in larger rewards for correct answers and larger penalties for incorrect ones. We use the exponential transformation rather than linear functions of entropy to ensure the confidence score is bounded and emphasizes the difference between high and low confidence predictions.

The outcome reward is:
\begin{equation}
\label{eq:outcome_reward}
r_{\text{outcome}}(q, y) = \begin{cases}
r_+ & \text{if extracted answer is correct} \\
r_- & \text{otherwise}
\end{cases}
\end{equation}
where $r_+ > 0$ and $r_- < 0$. The total reward combines both components:
\begin{equation}
r(q, y) = r_{\text{outcome}}(q, y) + r_{\text{confidence}}(q, y)
\end{equation}
\label{eq:total_reward}

\subsubsection{Optimization with GRPO}
We optimize the policy using Group Relative Policy Optimization (GRPO)~\citep{deepseek_math}, which eliminates the need for a separately trained value network by estimating advantages from a group of sampled trajectories. Our training setup involves two models: an \textbf{actor} $\pi_\theta$ initialized from the CoMT-tuned model that generates reasoning trajectories, and a \textbf{reference} model $\pi_{\text{ref}}$, a frozen copy of the CoMT-tuned model used for KL divergence regularization.

For each problem $q$ in a training batch, we sample a group of $G$ trajectories $\{y_i\}_{i=1}^{G}$ from the old policy $\pi_{\theta_{\text{old}}}$ and compute their rewards $\{r(q, y_i)\}_{i=1}^{G}$ as defined in Section \ref{eq:total_reward}. The advantage of each trajectory is then estimated by normalizing rewards within the group, and the policy is updated via a clipped surrogate objective with a KL penalty against the reference model. Full algorithm details are provided in Appendix~\ref{appendix:grpo_details}.

\begin{table*}[h]
\caption{In-distribution performance comparison across different models and methods on GSM8K and SVAMP benchmarks. Zero-shot and few-shot prompting baselines evaluate base model capabilities without fine-tuning. DeepSeek-Math and Qwen2.5-Math are specialized mathematical reasoning models included for reference. CoT-SFT, CoMT, CoT-SFT+RL, and CoMT+CCRL are trained on GSM8K and SVAMP training data. Cell shading compares against CoT-SFT+RL: \colorbox{green!20}{Green cells} indicates improvement, \colorbox{red!20}{Red cells} indicates degradation.}
  \label{tab:in-domain}
  \begin{center}
    \begin{small}
      \begin{sc}
        \setlength{\aboverulesep}{0pt}
        \setlength{\belowrulesep}{0pt}
        \setlength{\extrarowheight}{.75ex}
      \resizebox{\textwidth}{!}{%
        \begin{tabular}{l|cc|cc|cc|cc|c}
          \toprule
          \multirow{2}{*}{Method} & \multicolumn{2}{c|}{LLaMA3.1-8B} & \multicolumn{2}{c|}{Qwen2.5-7B} & \multicolumn{2}{c|}{Qwen3-4B} & \multicolumn{2}{c|}{Qwen3-8B} & \multirow{2}{*}{Avg.} \\
          \cmidrule(lr){2-3} \cmidrule(lr){4-5} \cmidrule(lr){6-7} \cmidrule(lr){8-9}
          & GSM8K & SVAMP & GSM8K & SVAMP & GSM8K & SVAMP & GSM8K & SVAMP & \\
          \midrule
          Zero-Shot                        & 62.90 & 67.60 & 72.60 & 83.00 & 84.70 & 85.00 & 70.10 & 64.40 & 73.79 \\
          Few-Shot                         & 77.50 & 84.00 & 90.10 & 92.20 & 84.80 & 91.60 & 90.00 & 93.10 & 87.91 \\
          \hline
          DeepSeek-Math  & \cellcolor{mathcolor}78.70 & \cellcolor{mathcolor}82.20 & \cellcolor{mathcolor}78.70 & \cellcolor{mathcolor}82.20 & \cellcolor{mathcolor}78.70 & \cellcolor{mathcolor}82.20 & 
          \cellcolor{mathcolor}78.70 & \cellcolor{mathcolor}82.20 & \cellcolor{mathcolor}80.45 \\
          Qwen2.5-Math & \cellcolor{mathcolor}79.20 & \cellcolor{mathcolor}85.50 & \cellcolor{mathcolor}79.20 & \cellcolor{mathcolor}85.50 & \cellcolor{mathcolor}79.20 & \cellcolor{mathcolor}85.50 & 
          \cellcolor{mathcolor}79.20 & \cellcolor{mathcolor}85.50 &
          \cellcolor{mathcolor}82.35 \\
          \hline
          CoT-SFT                 & \cellcolor{red!20}80.50 & \cellcolor{red!20}79.20 & \cellcolor{red!20}77.60 & \cellcolor{green!20}87.90 & \cellcolor{red!20}87.20 & \cellcolor{red!20}86.70 & \cellcolor{red!20}87.30 & \cellcolor{red!20}81.60 & \cellcolor{red!20}83.50 \\
          CoMT   & \cellcolor{green!20}82.80 & \cellcolor{green!20}83.10 & \cellcolor{green!20}86.70 & \cellcolor{red!20}87.30 & \cellcolor{green!20}89.30 & \cellcolor{green!20}90.10 & \cellcolor{green!20}88.80 & \cellcolor{green!20}91.20 & \cellcolor{green!20}87.41            \\ 
          
          \hline
          CoT-SFT+RL        & \cellcolor{red!20}86.30 & \cellcolor{red!20}91.60 & \cellcolor{red!20}87.60 & \cellcolor{red!20}92.60 & \cellcolor{red!20}86.70 & \cellcolor{red!20}89.00 & \cellcolor{red!20}89.50 & \cellcolor{red!20}85.70 & \cellcolor{red!20}88.62\\
          CoMT+CCRL & \cellcolor{green!20}86.70 & \cellcolor{green!20}92.40 & \cellcolor{green!20}88.90 & \cellcolor{green!20}93.10 & \cellcolor{green!20}91.60 & \cellcolor{green!20}91.50 & \cellcolor{green!20}91.00
          & \cellcolor{green!20}90.60 & \cellcolor{green!20}90.72 \\
          \bottomrule
        \end{tabular}
        }
      \end{sc}
    \end{small}
  \end{center}
  \vskip -0.2in
\end{table*}

\section{Experiments}
\subsection{Experimental Setup}
\label{sec:exp_setup}
\textbf{Datasets:} We evaluate our framework on multiple mathematical reasoning benchmarks to assess both in-distribution and out-of-distribution performance. For \textbf{in-distribution evaluation}, we use GSM8K~\citep{gsm8k} and SVAMP~\citep{svamp}. For \textbf{out-of-distribution evaluation}, we test on four additional datasets: AsDiv~\citep{asdiv}, MAWPS~\citep{mawps} , TabMWP~\citep{tabmwp}, GSM-Hard~\citep{gsm_hard}. 
For comprehensive evaluation, we also use GSM-Symbolic~\cite{gsm_symbolic} with symbolic perturbations and three hard datasets AIME25, AMC23 and Minerva-Math~\cite{minerva-math} in our experiments.

\textbf{Models:} For \textbf{training data generation}, we use large teacher model: LLaMA-3.1-70B-Instruct~\citep{llama3.1}. For \textbf{target models}, we fine-tune multiple checkpoints: LLaMA-3.1-8B-Instruct~\citep{llama3.1}, Qwen2.5-7B-Instruct~\citep{qwen2}, Qwen3-4B-Instruct and Qwen3-8B~\citep{qwen3}.

\textbf{Baselines:} We compare Ours against the following baselines: \textbf{Base Model}: The original model without additional mathematical reasoning training. \textbf{Math-Specialized Model}: Representative models fine-tuned specifically for mathematical reasoning at comparable scale, including DeepSeek-Math-7B-Instruct~\citep{deepseek_math} and Qwen2.5-Math-7B~\citep{qwen2.5_math}. \textbf{CoT-SFT+RL Model}: Models fine-tuned on complete chain-of-thought solution trajectories with specific calculations, followed by outcome-based reinforcement learning. This represents the current mainstream post-training approach.


\textbf{Training Details:} For both CoT-SFT+RL and CoMT+CCRL, we use the complete training sets of GSM8K and SVAMP as our base datasets. For CoT-SFT, we collect complete chain-of-thought solution trajectories for these problems using the teacher model (LLaMA-3.1-70B-Instruct). For CoMT, we generate meta-thought trajectories following the prompting strategy described in Section~\ref{comt_data_generation}. During the RL phase, both baselines and our method use the same hyperparameters for fair comparison

\textbf{Evaluation Protocol.} We evaluate all models using greedy decoding (temperature=0) with the standard ``Let's think step by step'' prompting strategy. We report accuracy as the primary metric, where a prediction is considered correct if the final numerical answer matches the ground truth.

\subsection{Experimenal Results}
\label{sec:exp_results}
We evaluate CoMT+CCRL from four perspectives in the main text: in-distribution performance on GSM8K and SVAMP (Section~\ref{sec:in_dist}),  ablation study on confidence reward design (Section~\ref{sec:ablation}), robustness to teacher model selection (Section~\ref{sec:teacher_robustness}) and overconfidence analysis (Section~\ref{sec:overconfidence}).

We further provide a series of analyses in the appendix, organized around the claims they support. To assess \emph{generalization}, we evaluate out-of-distribution transfer abilities on five benchmarks (Appendix~\ref{appendix:ood}) and performance on harder mathematical problems (Appendix~\ref{appendix:harder_benchmarks}). To verify that the gains arise from genuine \emph{abstraction} rather than confounding factors, we compare CoMT against length-matched and rewritten CoT supervision (Appendix~\ref{appendix:abstraction_vs_length}), measure robustness to symbolic perturbations that preserve reasoning structure while altering surface formulation (Appendix~\ref{appendix:robustness}), and quantify whether the model applies consistent strategies across problem variants (Appendix~\ref{appendix:strategy_consistency}). To establish \emph{robustness} of the framework itself, we show the gains persist when replacing GRPO with PPO (Appendix~\ref{appendix:ppo_results}) and demonstrate that our enhanced 7B model rivals models 2--4.5$\times$ larger (Appendix~\ref{appendix:different_size}). Finally, we report \emph{efficiency and qualitative} analyses: training-time and token savings over standard CoT-SFT (Appendix~\ref{appendix:efficiency}), sensitivity to the confidence-reward coefficients (Appendix~\ref{appendix:hyperparameter}), and a case study illustrating how the two stages jointly correct reasoning failures (Appendix~\ref{appendix:case_study}).

\begin{table*}[t]
\caption{Ablation study on confidence measure design in CCRL. We compare three alternatives against our proposed $H_{\max}$ formulation: \emph{w/o $r_{\text{conf}}$} (no confidence reward, outcome-only), $H_{\min}$ (confidence at the most certain computed step), and $H_{\text{avg}}$ (averaged across all computed steps). All models are trained on GSM8K and evaluated under greedy decoding. Row shading: \colorbox{green!20}{green} indicates $H_{\max}$ (best); \colorbox{red!20}{red} indicates the other variants.}
  \label{tab:entropy_ablation}
  \begin{center}
    \begin{small}
      \begin{sc}
        \setlength{\aboverulesep}{0pt}
        \setlength{\belowrulesep}{0pt}
        \setlength{\extrarowheight}{.75ex}
        \resizebox{\textwidth}{!}{%
        \begin{tabular}{l|l|c|ccccc|c}
          \toprule
          Model & Method & GSM8K & AsDiv & SVAMP & MAWPS & TabMWP & GSM\_Hard & Avg. \\
          \midrule
          \multirow{4}{*}{LLaMA3.1-8B}
          & w/o $r_{\text{conf}}$ & \cellcolor{red!20}85.20 & \cellcolor{red!20}86.50 & \cellcolor{red!20}84.70 & \cellcolor{red!20}95.40 & \cellcolor{red!20}60.00 & \cellcolor{red!20}43.90 & \cellcolor{red!20}75.95 \\
          & $H_{\min}$            & \cellcolor{red!20}84.10 & \cellcolor{red!20}85.20 & \cellcolor{red!20}85.50 & \cellcolor{red!20}95.00 & \cellcolor{red!20}60.40 & \cellcolor{red!20}42.60 & \cellcolor{red!20}75.47 \\
          & $H_{\text{avg}}$      & \cellcolor{red!20}86.00 & \cellcolor{red!20}83.50 & \cellcolor{red!20}85.40 & \cellcolor{red!20}95.40 & \cellcolor{red!20}59.90 & \cellcolor{red!20}42.40 & \cellcolor{red!20}75.43 \\
          \cline{2-9}
          & $H_{\max}$ (Ours)     & \cellcolor{green!20}86.70 & \cellcolor{green!20}87.90 & \cellcolor{green!20}87.00 & \cellcolor{green!20}96.20 & \cellcolor{green!20}61.10 & \cellcolor{green!20}45.10 & \cellcolor{green!20}77.33 \\
          \midrule
          \multirow{4}{*}{Qwen2.5-7B}
          & w/o $r_{\text{conf}}$ & \cellcolor{red!20}88.30 & \cellcolor{red!20}90.50 & \cellcolor{red!20}90.40 & \cellcolor{red!20}96.37 & \cellcolor{red!20}64.50 & \cellcolor{red!20}63.80 & \cellcolor{red!20}82.31 \\
          & $H_{\min}$            & \cellcolor{red!20}88.20 & \cellcolor{red!20}90.30 & \cellcolor{red!20}89.40 & \cellcolor{red!20}96.80 & \cellcolor{red!20}63.50 & \cellcolor{red!20}63.70 & \cellcolor{red!20}81.98 \\
          & $H_{\text{avg}}$      & \cellcolor{red!20}88.10 & \cellcolor{red!20}90.40 & \cellcolor{red!20}90.10 & \cellcolor{red!20}96.60 & \cellcolor{red!20}62.70 & \cellcolor{red!20}63.80 & \cellcolor{red!20}81.95 \\
          \cline{2-9}
          & $H_{\max}$ (Ours)     & \cellcolor{green!20}88.90 & \cellcolor{green!20}92.50 & \cellcolor{green!20}90.60 & \cellcolor{green!20}97.00 & \cellcolor{green!20}73.90 & \cellcolor{green!20}64.10 & \cellcolor{green!20}84.50 \\
          \midrule
          \multirow{4}{*}{Qwen3-4B}
          & w/o $r_{\text{conf}}$ & \cellcolor{red!20}90.90 & \cellcolor{red!20}90.70 & \cellcolor{red!20}91.80 & \cellcolor{red!20}96.40 & \cellcolor{red!20}71.40 & \cellcolor{red!20}61.90 & \cellcolor{red!20}83.85 \\
          & $H_{\min}$            & \cellcolor{red!20}91.10 & \cellcolor{red!20}90.90 & \cellcolor{red!20}91.20 & \cellcolor{red!20}96.20 & \cellcolor{red!20}71.20 & \cellcolor{red!20}61.40 & \cellcolor{red!20}83.67 \\
          & $H_{\text{avg}}$      & \cellcolor{red!20}91.20 & \cellcolor{red!20}89.60 & \cellcolor{red!20}91.60 & \cellcolor{red!20}96.00 & \cellcolor{red!20}70.20 & \cellcolor{red!20}61.50 & \cellcolor{red!20}83.35 \\
          \cline{2-9}
          & $H_{\max}$ (Ours)     & \cellcolor{green!20}91.60 & \cellcolor{green!20}91.20 & \cellcolor{green!20}93.30 & \cellcolor{green!20}96.70 & \cellcolor{green!20}72.60 & \cellcolor{green!20}62.90 & \cellcolor{green!20}84.72 \\
          \midrule
          \multirow{4}{*}{Qwen3-8B}
          & w/o $r_{\text{conf}}$ & \cellcolor{red!20}90.40 & \cellcolor{red!20}90.20 & \cellcolor{red!20}89.70 & \cellcolor{red!20}95.20 & \cellcolor{red!20}72.60 & \cellcolor{red!20}63.70 & \cellcolor{red!20}83.63 \\
          & $H_{\min}$            & \cellcolor{red!20}90.00 & \cellcolor{red!20}89.80 & \cellcolor{red!20}90.90 & \cellcolor{red!20}95.90 & \cellcolor{red!20}73.20 & \cellcolor{red!20}64.00 & \cellcolor{red!20}83.97 \\
          & $H_{\text{avg}}$      & \cellcolor{red!20}90.70 & \cellcolor{red!20}90.60 & \cellcolor{red!20}89.90 & \cellcolor{red!20}95.30 & \cellcolor{red!20}72.90 & \cellcolor{red!20}63.90 & \cellcolor{red!20}83.88 \\
          \cline{2-9}
          & $H_{\max}$ (Ours)     & \cellcolor{green!20}91.00 & \cellcolor{green!20}91.00 & \cellcolor{green!20}90.90 & \cellcolor{green!20}96.10 & \cellcolor{green!20}73.60 & \cellcolor{green!20}64.40 & \cellcolor{green!20}84.50 \\
          \bottomrule
        \end{tabular}%
        }
      \end{sc}
    \end{small}
  \end{center}
  \vskip -0.2in
\end{table*}

\subsubsection{In-Distribution Results}
\label{sec:in_dist}
Table~\ref{tab:in-domain} presents in-distribution performance comparison on GSM8K and SVAMP benchmarks across four base models, comparing our approach (CoMT+CCRL) against standard CoT-SFT+RL baselines and specialized mathematical reasoning models (DeepSeek-Math and Qwen2.5-Math).

\noindent\textbf{CoMT+CCRL substantially outperforms all baselines.} Our full pipeline achieves an average accuracy of 90.72\%, surpassing the standard CoT-SFT+RL baseline by +2.10 points (88.62\%) and significantly outperforming specialized math models: +10.27 points over DeepSeek-Math (80.45\%) and +8.38 points over Qwen2.5-Math (82.35\%). These improvements are consistent across different model sizes and architectures, demonstrating the effectiveness of our cognitively-aligned framework that separates meta-knowledge acquisition from task adaptation.

\noindent\textbf{Both CoMT and CCRL contribute substantially to overall gains.} Decomposing the improvements reveals complementary benefits from each component. CoMT alone outperforms standard CoT-SFT by +3.91 points (87.41\% vs.\ 83.50\%) and surpasses specialized math models by +6.96 and +5.06 points respectively, validating that focusing on abstract problem-solving patterns enables more effective meta-knowledge acquisition than learning from complete solution trajectories. Subsequently, adding CCRL provides further gains of +3.31 points (90.72\% vs.\ 87.41\%), demonstrating that confidence-based calibration effectively enhances task adaptation by preventing overconfident errors from cascading through reasoning steps.

\begin{figure*}[t]
  \centering
  \begin{subfigure}{\columnwidth}
    \centering
    \includegraphics[width=\columnwidth]{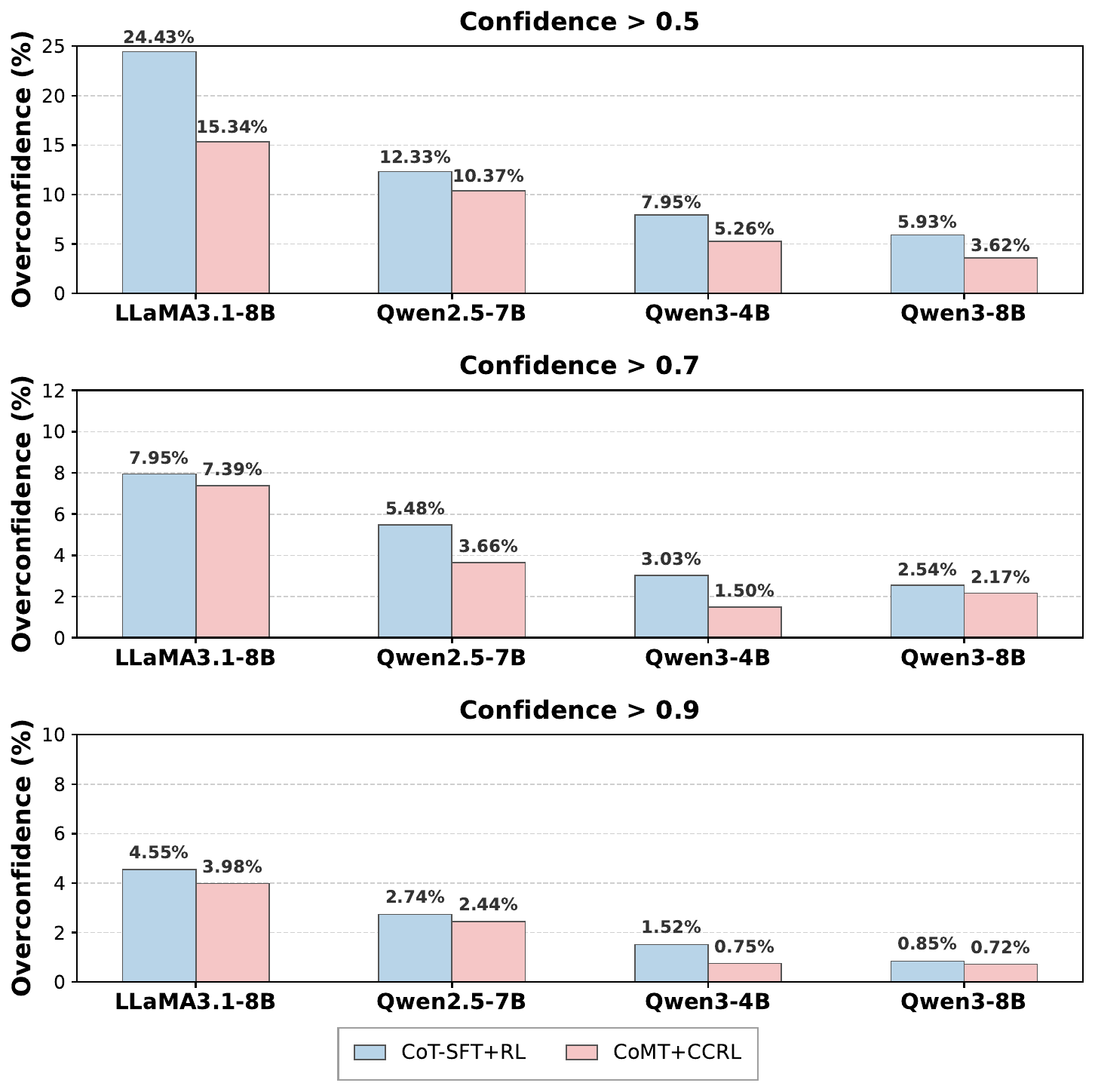}
    \caption{High-confidence errors on \textit{incorrect} predictions.}
    \label{fig:overconfidence_incorrect}
  \end{subfigure}
  \begin{subfigure}{\columnwidth}
    \centering
    \includegraphics[width=\columnwidth]{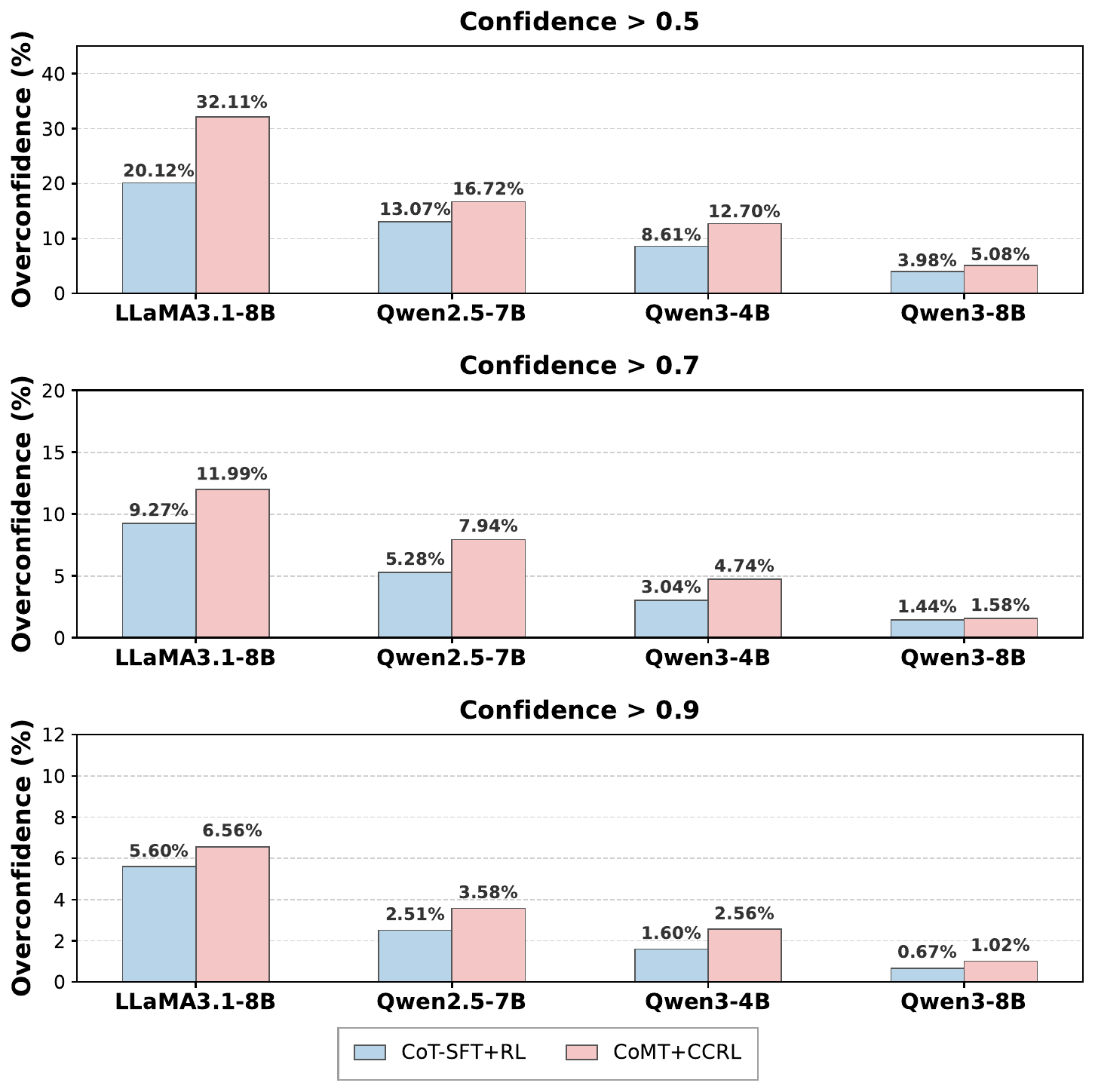}
    \caption{High-confidence rates on \textit{correct} predictions.}
    \label{fig:overconfidence_correct}
  \end{subfigure}
  \caption{Confidence calibration analysis at three thresholds ($c>0.5$, $0.7$, $0.9$), with confidence computed as in Section~\ref{sec:ccrl_reward}. CoMT+CCRL (a) substantially reduces high-confidence errors on incorrect predictions and (b) increases the rate of high-confidence outputs on correct predictions, showing that CCRL teaches the model to be confident when correct and uncertain when wrong.}
  \label{fig:confidence_calibration}
  \vskip -0.2in
\end{figure*}

\subsubsection{Ablation Study}
\label{sec:ablation}
To isolate the contribution of CCRL and validate the design of our confidence measure, we conduct an ablation study comparing three alternative formulations on top of CoMT-tuned models: \emph{w/o $r_{\text{conf}}$} (no confidence reward, outcome-only), $H_{\min}$ (confidence at the most certain computed step), and $H_{\text{avg}}$ (averaged across all computed steps), against our proposed $H_{\max}$ formulation. All variants are evaluated across four base models on GSM8K and five OOD benchmarks (Table~\ref{tab:entropy_ablation}).

\noindent\textbf{$H_{\max}$ consistently outperforms all alternatives.} Our formulation achieves the best average performance on all four models, with gains of +1.38, +2.19, +0.87, and +0.87 points over the \emph{w/o $r_{\text{conf}}$} baseline. Moreover, both $H_{\min}$ and $H_{\text{avg}}$ variants \textit{underperform} the \emph{w/o $r_{\text{conf}}$} baseline on LLaMA3.1-8B, Qwen2.5-7B, and Qwen3-4B, indicating that naively adding any entropy-based signal is not only insufficient but can be harmful. This validates that the choice of confidence measure is critical: only by targeting the \textit{least confident} computed step ($H_{\max}$) can the reward effectively prevent overconfident errors from cascading, whereas signals dominated by confident steps ($H_{\min}$) or diluted across all steps ($H_{\text{avg}}$) fail to provide useful calibration pressure.

\subsubsection{Robustness to Teacher Selection}
\label{sec:teacher_robustness}

To examine whether our framework's effectiveness depends on the specific teacher model, we replace LLaMA-3.1-70B-Instruct~\cite{llama3.1} with DeepSeek-R1~\cite{deepseek_r1} as the teacher, keeping all other settings fixed. We use LLaMA-3.1-8B-Instruct and Qwen2.5-7B-Instruct as the target models, trained on GSM8K, and evaluated under greedy decoding.
\begin{table}[h]
\centering
\caption{Performance on GSM8K with DeepSeek-R1 as the teacher model.}
\label{tab:teacher_robustness}
\setlength{\aboverulesep}{0pt}
\setlength{\belowrulesep}{0pt}
\setlength{\extrarowheight}{.75ex}
\resizebox{\columnwidth}{!}{%
\begin{tabular}{l|cc|cc}
\toprule
Model & SFT & CoMT & SFT+RL & CoMT+CCRL \\
\midrule
LLaMA& \cellcolor{red!20}81.40 & \cellcolor{green!20}86.00 & \cellcolor{red!20}84.60 & \cellcolor{green!20}{86.80} \\
Qwen & \cellcolor{red!20}87.30 & \cellcolor{green!20}90.40 & \cellcolor{red!20}88.80 & \cellcolor{green!20}{90.80} \\
\bottomrule
\end{tabular}%
}
\vskip -0.2in
\end{table}

As shown in Table~\ref{tab:teacher_robustness}, CoMT outperforms SFT by +4.6 and +3.1 points, and CoMT+CCRL outperforms SFT+RL by +2.2 and +2.0 points on the two target models. The consistent gains across a substantially different teacher confirm that the advantage of abstract meta-thought supervision is robust to teacher model choice, rather than being an artifact of a specific teacher's generation style.

\subsubsection{Overconfidence Analysis}
\label{sec:overconfidence}

To evaluate confidence calibration, we analyze the confidence of intermediate computational steps in incorrect predictions. Following Section~\ref{sec:ccrl_reward}, for each incorrect prediction, we compute the entropy $H_t$ for all computed numbers (Equation~\ref{entropy_for_all_tokens}), then use $H_{\max}$ (Equation~\ref{entropy_for_one_position}), the maximum entropy across steps, as the uncertainty measure. The confidence score is $c = e^{-H_{\max}}$ (Equation~\ref{eq:confidence_reward}). We report the proportion of predictions exceeding confidence thresholds $c > 0.5, 0.7, 0.9$ on both incorrect and correct predictions, capturing overconfident errors and confident correct answers respectively. 

Figure~\ref{fig:confidence_calibration} compares CoT-SFT+RL and CoMT+CCRL along both axes. \textbf{On incorrect predictions (Figure~\ref{fig:overconfidence_incorrect}), CoMT+CCRL reduces overconfidence across all models and thresholds.} At $c > 0.5$, high-confidence errors drop from 24.43\% to 15.34\% on LLaMA3.1-8B (37\% reduction) and from 12.33\% to 10.37\% on Qwen2.5-7B. \textbf{On correct predictions (Figure~\ref{fig:overconfidence_correct}), CoMT+CCRL instead \emph{increases} the rate of high-confidence outputs}, e.g., from 20.12\% to 32.11\% on LLaMA3.1-8B and from 13.07\% to 16.72\% on Qwen2.5-7B at $c > 0.5$. Together, these two trends confirm that CCRL produces well-calibrated confidence for reliable strategy execution: the model becomes more confident when it is right and more uncertain when it is wrong. 

\section{Conclusion}
We identified a misalignment between current post-training methods and human cognition: while humans decompose problem-solving into acquiring abstract strategies and adapting them to instances, existing methods optimize complete trajectories as atomic units, entangling strategy formation with execution and preventing meta-knowledge extraction. We proposed a cognitively-inspired framework with \textbf{C}hain-\textbf{o}f-\textbf{M}eta-\textbf{T}hought (\textbf{CoMT}) for learning abstract reasoning patterns and \textbf{C}onfidence-\textbf{C}alibrated \textbf{R}einforcement \textbf{L}earning (\textbf{CCRL}) for reliable task adaptation via confidence-aware rewards. Extensive experiments demonstrate that our framework consistently outperforms standard methods across multiple models and benchmarks, achieving 2.10\% and 3.86\% improvements in-distribution and out-of-distribution, respectively.

\bibliography{main}

\appendix


\clearpage
\onecolumn
\section*{Appendix Overview}
\label{sec:appendix-overview}

The appendices are ordered to build the argument in stages: after the
implementation details, we first establish \emph{generalization}; then verify that the gains stem from genuine \emph{abstraction} rather than confounding factors; next confirm \emph{robustness} across optimizers, scales, and perturbations; and finally report \emph{efficiency and qualitative} analyses.

\vspace{1.2em}

\newcommand{\appentry}[3]{%
  \item[#1]\,#2\dotfill p.~\pageref{#3}}

\noindent\textbf{Implementation}
\begin{description}[leftmargin=2.2em,labelsep=0.6em,itemsep=0.4em,topsep=0.3em]
  \appentry{A}{Training Details of GRPO}{appendix:grpo_details}
\end{description}

\vspace{0.4em}
\noindent\textbf{Generalization}
\begin{description}[leftmargin=2.2em,labelsep=0.6em,itemsep=0.4em,topsep=0.3em]
  \appentry{B}{Out-of-distribution Evaluation}{appendix:ood}
  \appentry{C}{Results on Harder Mathematical Benchmarks}{appendix:harder_benchmarks}
\end{description}

\vspace{0.4em}
\noindent\textbf{Abstraction (mechanism)}
\begin{description}[leftmargin=2.2em,labelsep=0.6em,itemsep=0.4em,topsep=0.3em]
  \appentry{D}{Abstraction vs.\ Shorter or Rewritten Supervision}{appendix:abstraction_vs_length}
  \appentry{E}{Robustness to Symbolic Perturbations}{appendix:robustness}
  \appentry{F}{Strategy Consistency Analysis}{appendix:strategy_consistency}
\end{description}

\vspace{0.4em}
\noindent\textbf{Robustness}
\begin{description}[leftmargin=2.2em,labelsep=0.6em,itemsep=0.4em,topsep=0.3em]
  \appentry{G}{Results with PPO as RL Optimizer}{appendix:ppo_results}
  \appentry{H}{Performance vs.\ Model Size}{appendix:different_size}
\end{description}

\vspace{0.4em}
\noindent\textbf{Efficiency \& Qualitative Analysis}
\begin{description}[leftmargin=2.2em,labelsep=0.6em,itemsep=0.4em,topsep=0.3em]
  \appentry{I}{Training Efficiency}{appendix:efficiency}
  \appentry{J}{Parameter Sensitivity}{appendix:hyperparameter}
  \appentry{K}{Case Study}{appendix:case_study}
\end{description}

\twocolumn

\begin{table*}[t]
\caption{Out-of-distribution performance comparison across different models and methods. Models trained on GSM8K are evaluated on five diverse benchmarks to assess generalization: AsDiv, SVAMP, MAWPS, TabMWP, and GSM-Hard. Zero-shot and few-shot baselines evaluate base model capabilities without fine-tuning. DeepSeek-Math and Qwen2.5-Math are specialized mathematical reasoning models included for reference. Cell shading compares against CoT-SFT+RL: \colorbox{green!20}{Green cells} indicates improvement, \colorbox{red!20}{Red cells} indicates degradation.}
  \label{tab:ood}
  \begin{center}
    \begin{small}
      \begin{sc}
      \setlength{\aboverulesep}{0pt}
\setlength{\belowrulesep}{0pt}
\setlength{\extrarowheight}{.75ex}
        \begin{tabular}{l|l|ccccc|c}
          \toprule
          Model & Method & AsDiv & SVAMP & MAWPS & TabMWP & GSM\_Hard & Avg. \\
          \hline
          DeepSeek-Math & Zero-shot & \cellcolor{mathcolor}85.00 &\cellcolor{mathcolor}82.20	&\cellcolor{mathcolor}92.50	&\cellcolor{mathcolor}69.90	&\cellcolor{mathcolor}56.10 & \cellcolor{mathcolor}77.14 \\
          Qwen2.5-Math & Zero-shot & \cellcolor{mathcolor}82.50	&\cellcolor{mathcolor}85.50	&\cellcolor{mathcolor}92.30	&\cellcolor{mathcolor}53.60	&\cellcolor{mathcolor}55.40 & \cellcolor{mathcolor}73.86\\
          \midrule
          \multirow{6}{*}{LLaMA3.1-8B} 
          & zero-shot         & 63.60 & 67.60 & 73.40 & 39.50 & 31.70 & 55.16 \\
          & few-shot          & 85.80 & 84.00 & 97.00 & 55.00 & 38.70 & 72.10 \\
          \cline{2-8}
          & CoT-SFT           & \cellcolor{red!20}76.10 & \cellcolor{red!20}78.40 & \cellcolor{red!20}80.80 & \cellcolor{red!20}46.40 & \cellcolor{red!20}37.20 & \cellcolor{red!20}63.78 \\
          & CoMT       & \cellcolor{green!20}83.80 & \cellcolor{green!20}82.80 & \cellcolor{green!20}93.50 & \cellcolor{green!20}59.70 & \cellcolor{green!20}41.80 & \cellcolor{green!20}72.32 \\
          \cline{2-8}
          & CoT-SFT+RL        & \cellcolor{red!20}85.60 & \cellcolor{red!20}86.50 & \cellcolor{red!20}92.60 & \cellcolor{red!20}58.80 & \cellcolor{red!20}40.90 & \cellcolor{red!20}72.88 \\
          & CoMT+CCRL & \cellcolor{green!20}87.90 & \cellcolor{green!20}87.00 & \cellcolor{green!20}96.20 & \cellcolor{green!20}61.10 & \cellcolor{green!20}45.10 & \cellcolor{green!20}75.46 \\
          \midrule
          \multirow{6}{*}{Qwen2.5-7B} 
          & zero-shot         & 84.20 & 83.00 & 90.80 & 61.20 & 53.40 & 74.52 \\
          & few-shot          & 90.90 & 92.20 & 97.60 & 70.40 & 62.90 & 82.80 \\
          \cline{2-8}
          & CoT-SFT           & \cellcolor{red!20}81.60 & \cellcolor{red!20}86.80 & \cellcolor{red!20}85.70 & \cellcolor{red!20}53.30 & \cellcolor{red!20}52.80 & \cellcolor{red!20}72.04 \\
          & CoMT       & \cellcolor{green!20}89.60 & \cellcolor{green!20}89.10 & \cellcolor{green!20}96.80 & \cellcolor{green!20}61.60 & \cellcolor{green!20}63.40 & \cellcolor{green!20}80.10 \\
          \cline{2-8}
          & CoT-SFT+RL        & \cellcolor{red!20}90.10 & \cellcolor{red!20}90.50 & \cellcolor{red!20}96.10 & \cellcolor{red!20}63.10 & \cellcolor{red!20}60.40 & \cellcolor{red!20}80.04 \\
          & CoMT+CCRL & \cellcolor{green!20}92.50 & \cellcolor{green!20}90.60 & \cellcolor{green!20}97.00 & \cellcolor{green!20}73.90 & \cellcolor{green!20}64.10 & \cellcolor{green!20}83.62 \\
          \midrule
          \multirow{6}{*}{Qwen3-4B} 
          & zero-shot         & 78.10 & 85.00 & 88.10 & 66.30 & 62.40 & 75.98 \\
          & few-shot          & 88.40 & 91.60 & 95.90 & 70.30 & 56.00 & 80.44 \\
          \cline{2-8}
          & CoT-SFT           & \cellcolor{red!20}85.70 & \cellcolor{red!20}88.30 & \cellcolor{red!20}93.70 & \cellcolor{red!20}64.00 & \cellcolor{red!20}54.80 & \cellcolor{red!20}77.30 \\
          & CoMT       & \cellcolor{green!20}86.90 & \cellcolor{green!20}92.40 & \cellcolor{green!20}97.90 & \cellcolor{green!20}68.60 & \cellcolor{green!20}66.60 & \cellcolor{green!20}82.48 \\
          \cline{2-8}
          & CoT-SFT+RL        & \cellcolor{red!20}85.80 & \cellcolor{red!20}89.30 & \cellcolor{red!20}93.80 & \cellcolor{red!20}63.40 & \cellcolor{red!20}57.20 & \cellcolor{red!20}77.90 \\
          & CoMT+CCRL & \cellcolor{green!20}91.20 & \cellcolor{green!20}93.30 & \cellcolor{green!20}96.70 & \cellcolor{green!20}72.60 & \cellcolor{green!20}62.90 & \cellcolor{green!20}83.34 \\
          \midrule
          \multirow{6}{*}{Qwen3-8B} 
          & zero-shot         & 58.10 & 64.40 & 57.10 & 57.20 & 49.50 & 57.26 \\
          & few-shot          & 89.70 & 93.10 & 96.60 & 73.90 & 69.30 & 84.52 \\
          \cline{2-8}
          & CoT-SFT           & \cellcolor{red!20}81.60 & \cellcolor{red!20}81.90 & \cellcolor{red!20}84.70 & \cellcolor{red!20}69.70 & \cellcolor{red!20}53.90 & \cellcolor{red!20}74.36 \\
          & CoMT       & \cellcolor{green!20}87.60 & \cellcolor{green!20}89.70 & \cellcolor{green!20}95.80 & \cellcolor{green!20}74.30 & \cellcolor{green!20}62.50 & \cellcolor{green!20}81.98 \\
          \cline{2-8}
          & CoT-SFT+RL        & \cellcolor{red!20}85.60 & \cellcolor{red!20}87.80 & \cellcolor{red!20}89.20 & \cellcolor{red!20}71.20 & \cellcolor{red!20}63.00 & \cellcolor{red!20}79.36 \\
          & CoMT+CCRL & \cellcolor{green!20}91.00 & \cellcolor{green!20}90.90 & \cellcolor{green!20}96.10 & \cellcolor{green!20}73.60 & \cellcolor{green!20}64.40 & \cellcolor{green!20}83.20 \\
          \bottomrule
        \end{tabular}
      \end{sc}
    \end{small}
  \end{center}
  \vskip -0.1in
\end{table*}

\section{Training Details of GRPO}
\label{appendix:grpo_details}

We optimize the policy using Group Relative Policy Optimization (GRPO)~\citep{deepseek_math}. Unlike PPO~\citep{ppo}, GRPO eliminates the value network by estimating advantages from a group of sampled trajectories, reducing memory overhead while remaining effective for outcome-based rewards.

Our training setup involves two models: an \textbf{actor} $\pi_\theta$ initialized from the CoMT-tuned model, and a frozen \textbf{reference} model $\pi_{\text{ref}}$ for KL regularization.

For each problem $q$, we sample a group of $G$ trajectories $\{y_i\}_{i=1}^{G}$ from the old policy $\pi_{\theta_{\text{old}}}$ and compute their rewards $\{r_i\}_{i=1}^{G}$ following Eq.~\ref{eq:total_reward}. The advantage of each trajectory is estimated by normalizing rewards within the group:
\begin{equation}
\hat{A}_i = \frac{r_i - \mathrm{mean}(\{r_j\}_{j=1}^{G})}{\mathrm{std}(\{r_j\}_{j=1}^{G})}.
\end{equation}
Since rewards are assigned at the trajectory level, all tokens in $y_i$ share the same advantage $\hat{A}_i$.

Let $\rho_{i,t}(\theta) = \pi_\theta(y_{i,t} | q, y_{i,<t}) / \pi_{\theta_{\text{old}}}(y_{i,t} | q, y_{i,<t})$. The GRPO objective is:
\begin{equation}
\begin{aligned}
\mathcal{J}(\theta) = {E} \Bigg[ \frac{1}{G} \sum_{i=1}^{G} \frac{1}{T_i} \sum_{t=1}^{T_i} \Big( &\min\big(\rho_{i,t} \hat{A}_i,\ \\ \mathrm{clip}(\rho_{i,t}, 1-\epsilon, 1+\epsilon) \hat{A}_i\big) 
&- \beta\, {D}_{\text{KL}}\!\left[\pi_\theta \,\|\, \pi_{\text{ref}}\right] \Big) \Bigg],
\end{aligned}
\end{equation}
where $\epsilon$ is the clipping threshold and $\beta$ controls the KL penalty strength. Following~\citet{deepseek_math}, the KL term is added directly to the loss and estimated with the unbiased estimator of~\citet{approximating_kl}. We optimize with Adam~\citep{adam} and gradient clipping.

\section{Out-of-distribution Evaluation}
\label{appendix:ood}
Table~\ref{tab:ood} evaluates out-of-distribution generalization of models trained on GSM8K and SVAMP across five diverse mathematical reasoning benchmarks (AsDiv, MAWPS, TabMWP, and GSM-Hard, together with the held-out portion of SVAMP), comparing our approach against baselines.

\noindent\textbf{CoMT+CCRL demonstrates superior generalization across all settings.} Our full pipeline achieves an average accuracy of 81.41\%, substantially outperforming standard CoT-SFT+RL (77.55\%) by +3.86 points and surpassing specialized models DeepSeek-Math (77.14\%) and Qwen2.5-Math (73.86\%) by +4.27 and +7.55 points respectively. The improvements are consistent across all backbones, with per-model gains over CoT-SFT+RL ranging from +2.58 (LLaMA3.1-8B) to +5.44 (Qwen3-4B) points. Notably, our approach surpasses few-shot prompting baselines on average (81.41\% vs. 79.97\%) despite having no access to task-specific exemplars at inference time.

\noindent\textbf{Meta-knowledge acquisition drives stronger OOD generalization.} CoMT outperforms CoT-SFT by +7.35 points in OOD settings (79.22\% vs. 71.87\%), nearly double the +3.91 points improvement observed in-distribution (Table~\ref{tab:in-domain}). This amplified benefit demonstrates that abstract reasoning patterns provide substantially stronger generalization than trajectory-level imitation when facing distribution shifts. 


\section{Results on Harder Mathematical Benchmarks}
\label{appendix:harder_benchmarks}
\begin{table*}[h]
\caption{Performance on harder mathematical reasoning benchmarks. We evaluate CoT-SFT+RL and CoMT+CCRL across four backbones on AIME25, AMC23, and Minerva\_Math under both greedy decoding and pass@8 settings. All models are trained on GSM8K. Cell shading compares CoMT+CCRL against CoT-SFT+RL: \colorbox{green!20}{Green cells} indicate improvement, \colorbox{red!20}{Red cells} indicate degradation.}
  \label{tab:harder_benchmarks}
  \begin{center}
    \begin{small}
      \begin{sc}
      \setlength{\aboverulesep}{0pt}
\setlength{\belowrulesep}{0pt}
\setlength{\extrarowheight}{.75ex}
      \resizebox{\textwidth}{!}{%
        \begin{tabular}{l|l|cc|cc|cc|c}
          \toprule
          \multirow{2}{*}{Model} & \multirow{2}{*}{Method} & \multicolumn{2}{c|}{AIME25} & \multicolumn{2}{c|}{AMC23} & \multicolumn{2}{c|}{Minerva\_Math} & \multirow{2}{*}{Avg.} \\
          \cmidrule(lr){3-4} \cmidrule(lr){5-6} \cmidrule(lr){7-8}
          & & Greedy & Pass@8 & Greedy & Pass@8 & Greedy & Pass@8 & \\
          \midrule
          \multirow{2}{*}{LLaMA3.1-8B}
          & CoT-SFT+RL  & \cellcolor{red!20}0.00 & \cellcolor{red!20}3.33 & \cellcolor{red!20}22.50 & \cellcolor{red!20}62.50 & \cellcolor{red!20}27.57 & \cellcolor{red!20}52.57 & \cellcolor{red!20}28.08 \\
          & CoMT+CCRL   & \cellcolor{green!20}0.00 & \cellcolor{green!20}6.67 & \cellcolor{green!20}27.50 & \cellcolor{green!20}67.50 & \cellcolor{green!20}29.04 & \cellcolor{green!20}55.51 & \cellcolor{green!20}31.04 \\
          \midrule
          \multirow{2}{*}{Qwen2.5-7B}
          & CoT-SFT+RL  & \cellcolor{red!20}6.67 & \cellcolor{red!20}23.33 & \cellcolor{red!20}40.00 & \cellcolor{red!20}80.00 & \cellcolor{red!20}36.03 & \cellcolor{red!20}51.84 & \cellcolor{red!20}39.65 \\
          & CoMT+CCRL   & \cellcolor{green!20}10.00 & \cellcolor{green!20}26.67 & \cellcolor{green!20}50.00 & \cellcolor{green!20}80.00 & \cellcolor{green!20}39.34 & \cellcolor{green!20}54.04 & \cellcolor{green!20}43.34 \\
          \midrule
          \multirow{2}{*}{Qwen3-4B}
          & CoT-SFT+RL  & \cellcolor{red!20}33.33 & \cellcolor{red!20}66.67 & \cellcolor{red!20}87.50 & \cellcolor{red!20}95.00 & \cellcolor{red!20}40.81 & \cellcolor{red!20}52.21 & \cellcolor{red!20}62.59 \\
          & CoMT+CCRL   & \cellcolor{green!20}40.00 & \cellcolor{green!20}70.00 & \cellcolor{green!20}90.00 & \cellcolor{green!20}100.00 & \cellcolor{green!20}42.65 & \cellcolor{green!20}52.57 & \cellcolor{green!20}65.87 \\
          \midrule
          \multirow{2}{*}{Qwen3-8B}
          & CoT-SFT+RL  & \cellcolor{red!20}50.00 & \cellcolor{red!20}73.33 & \cellcolor{red!20}85.00 &\cellcolor{red!20}100.00 & \cellcolor{red!20}41.91 & \cellcolor{red!20}55.14 & \cellcolor{red!20}67.56 \\
          & CoMT+CCRL   & \cellcolor{green!20}56.67 & \cellcolor{green!20}76.67 & \cellcolor{green!20}92.50 & \cellcolor{green!20}100.00 & \cellcolor{green!20}44.85 & \cellcolor{green!20}55.88 & \cellcolor{green!20}71.09 \\
          \bottomrule
        \end{tabular}
        }
      \end{sc}
    \end{small}
  \end{center}
  \vskip -0.1in
\end{table*}

We evaluate CoT-SFT+RL and CoMT+CCRL on three substantially harder mathematical reasoning benchmarks: AIME25, AMC23, and Minerva\_Math~\citep{minerva-math}. All models are trained only on GSM8K. We report both greedy decoding accuracy and pass@8, the latter reflecting reasoning capability under sampling. As shown in Table~\ref{tab:harder_benchmarks}, CoMT+CCRL consistently outperforms CoT-SFT+RL across all four backbones, with average gains of +2.96 (LLaMA3.1-8B), +3.70 (Qwen2.5-7B), +3.28 (Qwen3-4B), and +3.53 (Qwen3-8B) points. These results demonstrate that the abstract reasoning patterns learned via CoMT transfer to harder problem distributions, supporting our claim that meta-knowledge acquisition produces robust generalization rather than benchmark-specific gains.

\section{Abstraction vs. Shorter or Rewritten Supervision}
\label{appendix:abstraction_vs_length}

To test whether CoMT's improvements stem from genuine abstraction of reasoning patterns or from confounding factors such as shorter supervision length or stylistic regularization of CoT trajectories, we construct two controlled baselines using DeepSeek-R1 as the teacher:
\begin{itemize}[itemsep=2pt, topsep=2pt, parsep=0pt, partopsep=0pt, leftmargin=*]
    \item \textbf{S-CoT}: The teacher generates CoT solutions under an explicit length constraint matching the average length of CoMT trajectories ($\sim$133 tokens on GSM8K), while preserving all specific numerical computations. This controls for supervision length.
    \item \textbf{R-CoT}: The teacher rewrites full CoT solutions as concisely as possible while retaining all numerical calculations, without explicit length constraints. This controls for stylistic regularization.
\end{itemize}
The key distinction between both baselines and CoMT is that they preserve concrete numerical computations, whereas CoMT explicitly replaces all specific numbers with variable names to enforce abstract reasoning patterns. For a fair comparison, we use DeepSeek-R1 as the teacher across all four variants, including our CoMT and CoMT+CCRL, and train on GSM8K under greedy decoding.

\textbf{Gains stem from abstraction, not from length or stylistic regularization.} As shown in Table~\ref{tab:abstraction_vs_length}, CoMT outperforms Short-CoT by +7.5 and +3.8 points and Rewrite-CoT by +13.4 and +8.6 points on the two target models, despite all three formats using the same teacher. Notably, both length-matched and rewritten baselines perform \textit{worse} than standard SFT with full CoT, suggesting that simply shortening or stylistically normalizing supervision is not only unhelpful but can degrade performance. In contrast, CoMT's replacement of numerical values with variable names yields substantial gains, providing direct evidence that the benefit comes from supervising abstract reasoning patterns rather than from supervision compression and confirming that abstraction-based supervision provides a stronger foundation for subsequent confidence-calibrated optimization.

\begin{table}[h]
\centering
\small
\caption{Comparison against length-matched (Short-CoT) and rewritten (Rewrite-CoT) baselines on GSM8K. All variants use DeepSeek-R1 as the teacher model.}
\label{tab:abstraction_vs_length}
\setlength{\aboverulesep}{0pt}
\setlength{\belowrulesep}{0pt}
\setlength{\extrarowheight}{.75ex}
\resizebox{\columnwidth}{!}{%
\begin{tabular}{l|cc}
\toprule
Method & LLaMA3.1-8B & Qwen2.5-7B \\
\midrule
S-CoT SFT       & \cellcolor{red!20}78.50 & \cellcolor{red!20}86.60 \\
R-CoT SFT     & \cellcolor{red!20}72.60 & \cellcolor{red!20}81.80 \\
SFT                 & \cellcolor{red!20}81.40 & \cellcolor{red!20}87.30 \\
CoMT                & \cellcolor{green!20}{86.00} & \cellcolor{green!20}{90.40} \\
\midrule
S-CoT SFT+RL    & \cellcolor{red!20}83.50 & \cellcolor{red!20}89.30 \\
R-CoT SFT+RL  & \cellcolor{red!20}82.20 & \cellcolor{red!20}84.60 \\
SFT+RL              & \cellcolor{red!20}84.60 & \cellcolor{red!20}88.80 \\
CoMT+CCRL           & \cellcolor{green!20}{86.80} & \cellcolor{green!20}{90.80} \\
\bottomrule
\end{tabular}
}
\end{table}

\section{Robustness to Symbolic Perturbations}
\label{appendix:robustness}

\begin{figure}[ht]
  \vskip 0.2in
  \begin{center}
    \centerline{\includegraphics[width=\columnwidth]{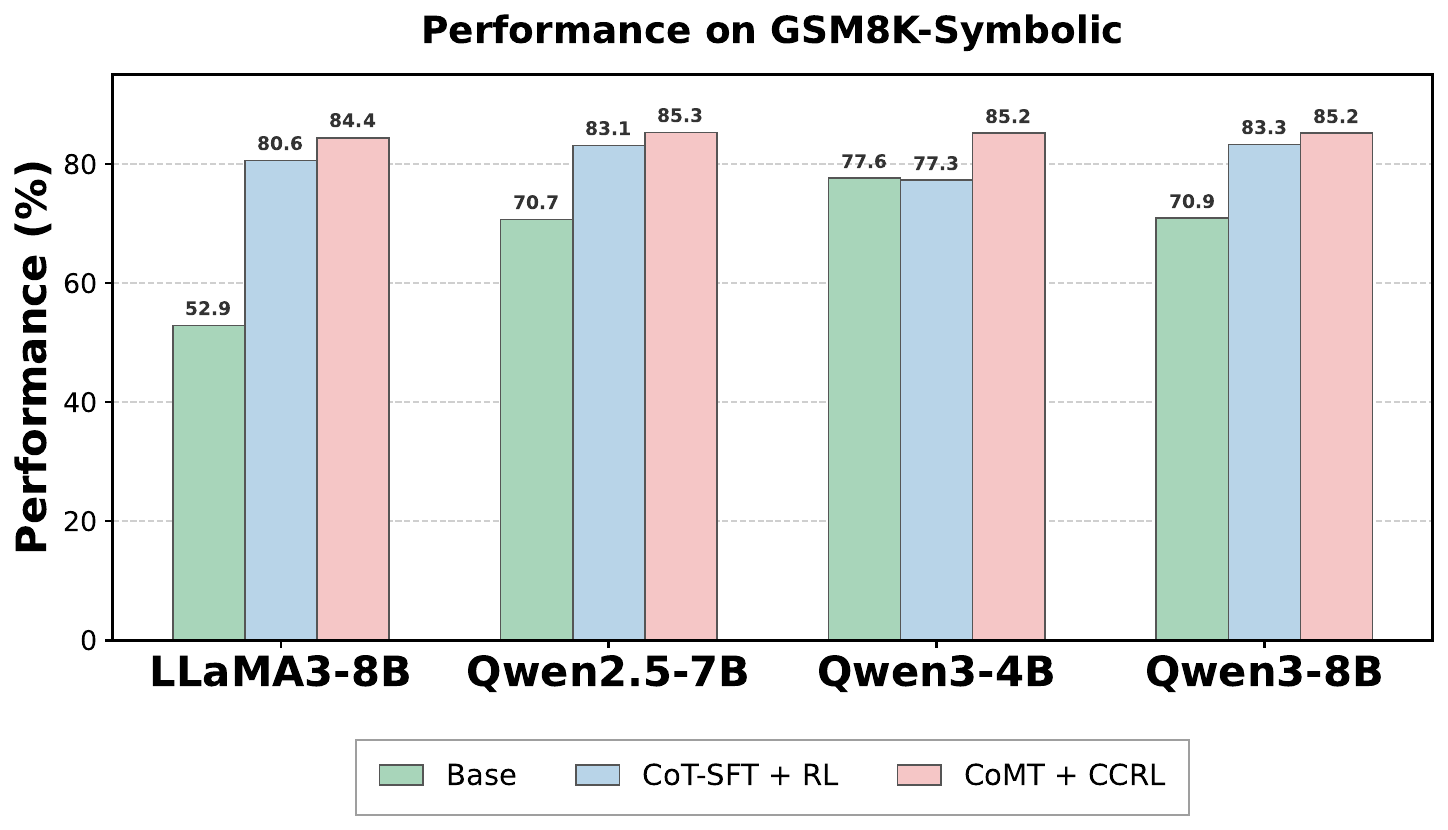}}
\caption{Generalization evaluation on GSM-Symbolic benchmark. GSM-Symbolic tests genuine reasoning capabilities through symbolic variations of mathematical problems. Models are trained on GSM8K and SVAMP, then evaluated on GSM-Symbolic. CoMT+CCRL consistently outperforms CoT-SFT+RL consistently across all models, suggesting that our framework enables more robust reasoning.}
    \label{fig:gsm_symbolic}
  \end{center}
     \vskip -0.4in
\end{figure}

To further validate the effectiveness of meta-knowledge acquisition, we evaluate on GSM-Symbolic~\citep{gsm_symbolic}, a benchmark specifically designed to test genuine reasoning through symbolic variations of mathematical problems. GSM-Symbolic requires models to apply abstract problem-solving strategies across problems with similar underlying reasoning patterns, making it an ideal benchmark for evaluating the meta-knowledge that our framework is designed to acquire.

\textbf{CoMT+CCRL demonstrates superior robustness through meta-knowledge acquisition.} Figure~\ref{fig:gsm_symbolic} shows that while all models experience performance drops compared to standard GSM8K, CoMT+CCRL maintains substantially stronger performance than CoT-SFT+RL. For instance, improvements of +3.8\% on LLaMA3-8B (80.6\% vs 84.4\%) and +2.2\% on Qwen2.5-7B (83.1\% vs 85.3\%). These consistent improvements validate our method: by explicitly learning abstract reasoning patterns rather than problem-specific solutions, models acquire meta-knowledge that generalizes to symbolic variations, thereby achieving robust generalization.

\section{Strategy Consistency Analysis}
\label{appendix:strategy_consistency}

While Section~\ref{appendix:robustness} shows that CoMT+CCRL achieves higher accuracy on GSM-Symbolic, accuracy alone does not reveal whether models apply consistent abstract strategies across variants or solve each via unrelated reasoning paths. To probe this, we conduct a strategy consistency analysis on GSM-Symbolic, where each group of 50 variants shares the same underlying reasoning structure but differs in surface numbers and entity names (\textit{i.e.}, formulations). For each group, we use GPT-4o~\cite{gpt4o} as a judge to cluster generated trajectories into distinct strategies, and report the percentage of variants following the dominant one: a higher score indicates that the model applies the same abstract strategy regardless of surface changes.

\begin{figure}[h]
  \centering
  \includegraphics[width=\columnwidth]{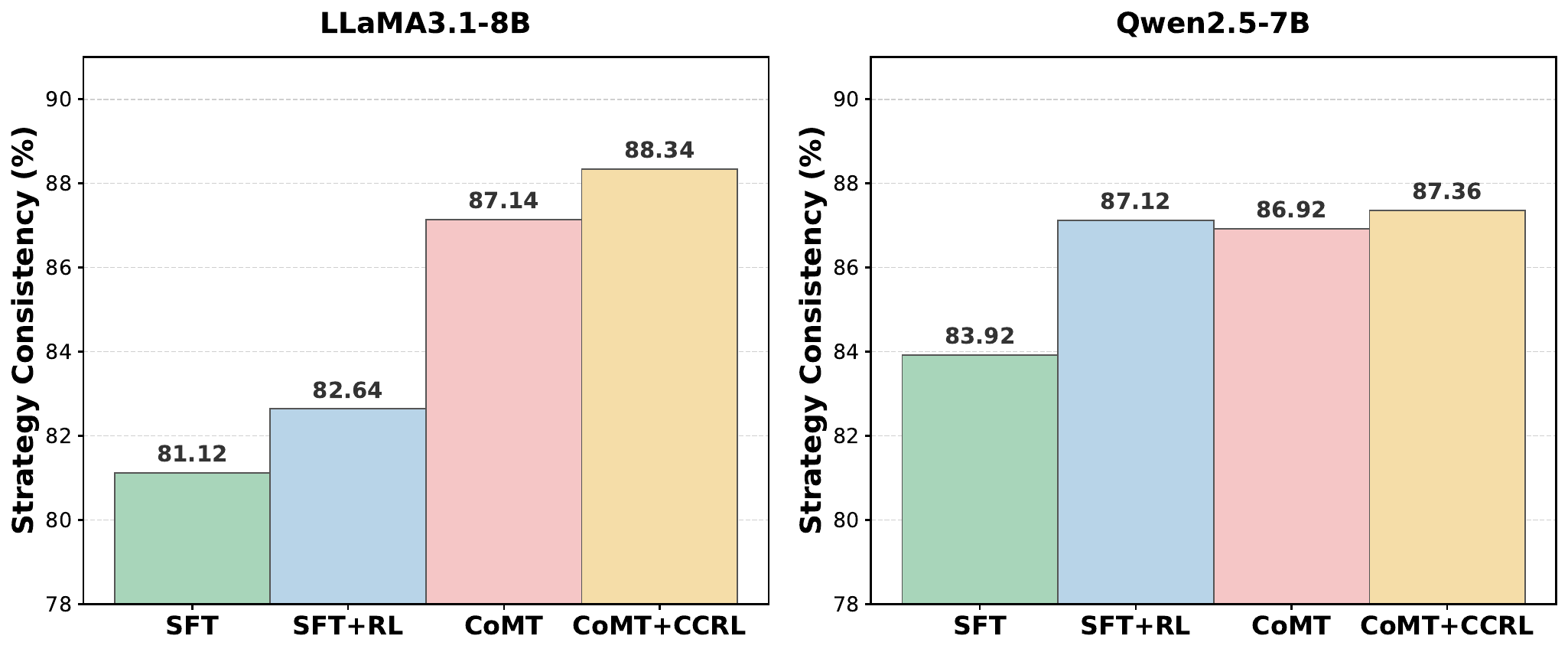}
  \caption{Strategy consistency on GSM-Symbolic. CoMT-based methods apply substantially more consistent strategies across variants sharing the same reasoning structure.}
  \label{fig:strategy_consistency}
\end{figure}

As shown in Figure~\ref{fig:strategy_consistency}, CoMT alone already improves consistency over SFT+RL (+4.5 points on LLaMA3.1-8B; comparable on Qwen2.5-7B), indicating that abstract supervision itself encourages transferable reasoning patterns rather than surface-level memorization. Adding CCRL further raises consistency to 88.34\% and 87.36\% respectively, validating the complementary design of our two stages: CoMT establishes strategy generalization, on which CCRL builds execution reliability without compromising it.

\section{Results with PPO as RL Optimizer}
\label{appendix:ppo_results}

To verify that the effectiveness of our framework is not tied to a specific RL algorithm, we re-run the pipeline with PPO~\citep{ppo} in place of GRPO, keeping all other settings identical. As shown in Table~\ref{tab:ppo_results}, CoMT+CCRL consistently outperforms CoT-SFT+RL on every benchmark across all four models, with average gains of +2.78 (LLaMA3.1-8B), +1.86 (Qwen2.5-7B), +6.00 (Qwen3-4B), and +6.00 (Qwen3-8B) points. The gains are particularly pronounced on harder OOD benchmarks such as TabMWP and GSM\_Hard (e.g., +13.2 points on GSM\_Hard for Qwen3-4B), mirroring the trends observed under GRPO and confirming that the benefits of our framework arise from the cognitively-inspired decomposition itself rather than from any specific policy optimization algorithm.

\section{Performance vs. Model Size}
\label{appendix:different_size}
\begin{figure}[t]
  \centering
  \includegraphics[width=0.95\columnwidth]{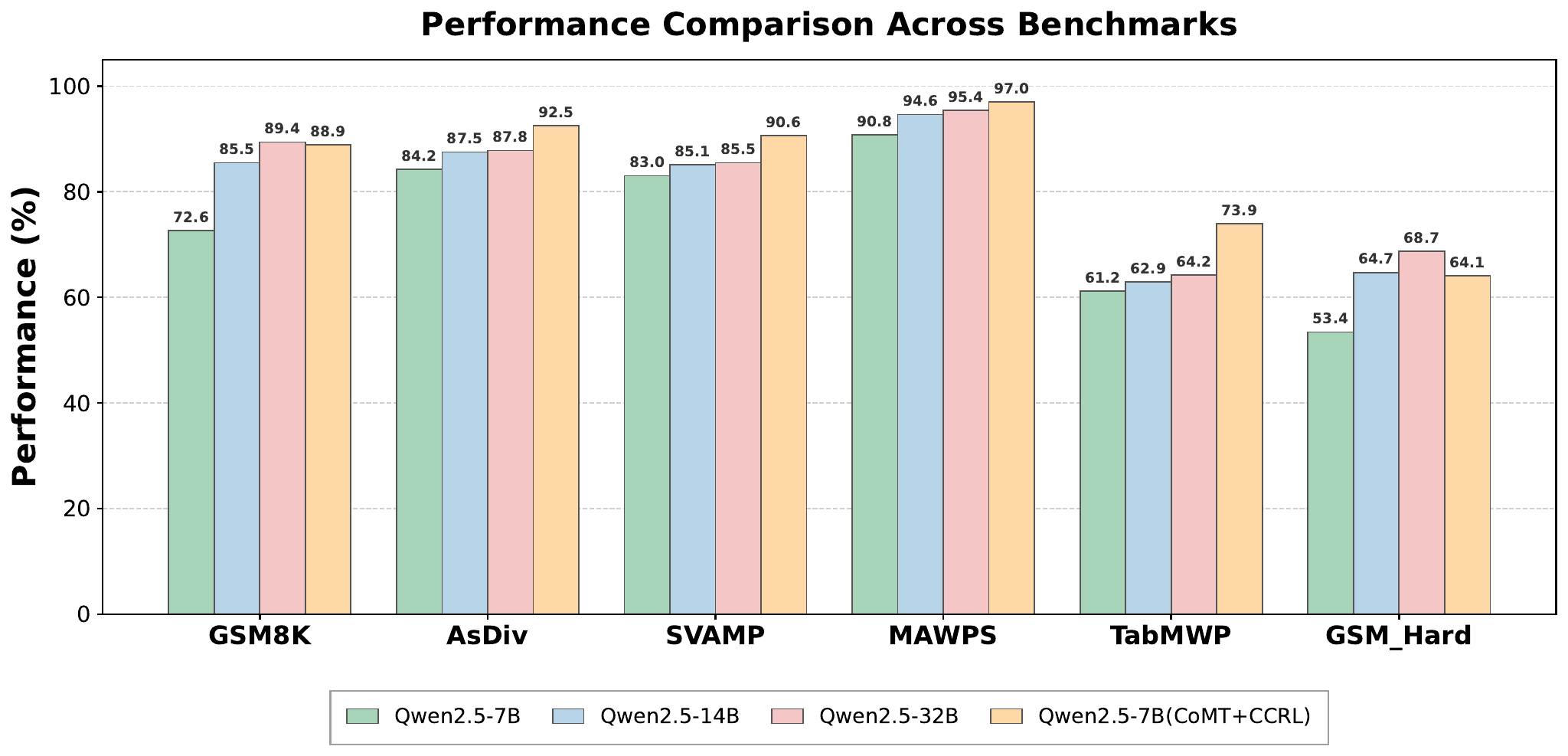}
 \caption{Performance comparison across mathematical reasoning benchmarks. Our method (CoMT+CCRL) applied to Qwen2.5-7B achieves competitive or superior performance compared to larger base models (14B and 32B parameters) across six diverse benchmarks, demonstrating the effectiveness of our cognitively-inspired training approach in enhancing reasoning capabilities without scaling model size.}
  \label{fig:performance_comparison}
  \vskip -0.2in
\end{figure}

Figure~\ref{fig:performance_comparison} compares our CoMT+CCRL trained Qwen2.5-7B model against larger base models (14B and 32B) across six benchmarks. Our enhanced 7B model achieves substantial improvements over its base version, with gains ranging from 6.2\% to 16.3\% across different benchmarks, and remarkably matches or exceeds the performance of models 2-4.5$\times$ larger. 

Notably, on four key benchmarks: AsDiv, SVAMP, MAWPS, and TabMWP, our CoMT+CCRL trained 7B model outperforms the 32B base model despite using significantly fewer parameters. Across all six benchmarks, our enhanced 7B model achieves 84.5\% average accuracy, representing an 8.8-point improvement over the base 7B model (74.2\%), a 3.0-point gain over the 14B base model (80.1\%), and even a 1.2-point advantage over the 32B base model (81.8\%). 

These results demonstrate that cognitively-inspired training which explicitly learning abstract strategies with confidence-calibrated execution can be more effective than simply scaling model size by 2-4.5$\times$ for mathematical reasoning tasks, achieving comparable or superior performance with substantially lower computational requirements.

\begin{table*}[t]
\caption{Performance comparison with PPO as the RL optimization algorithm. We compare CoT-SFT+RL against our CoMT+CCRL framework across four models on GSM8K and five OOD benchmarks. All models are trained on GSM8K and evaluated under greedy decoding. Cell shading compares CoMT+CCRL against CoT-SFT+RL: \colorbox{green!20}{green} indicates improvement, \colorbox{red!20}{red} indicates degradation.}
  \label{tab:ppo_results}
  \begin{center}
    \begin{small}
      \begin{sc}
      \setlength{\aboverulesep}{0pt}
\setlength{\belowrulesep}{0pt}
\setlength{\extrarowheight}{.75ex}
        \resizebox{\textwidth}{!}{%
        \begin{tabular}{l|l|c|ccccc|c}
          \toprule
          Model & Method & GSM8K & AsDiv & SVAMP & MAWPS & TabMWP & GSM\_Hard & Avg. \\
          \midrule
          \multirow{2}{*}{LLaMA3.1-8B}
          & CoT-SFT+RL  & \cellcolor{red!20}84.20 & \cellcolor{red!20}82.50 & \cellcolor{red!20}84.10 & \cellcolor{red!20}88.40 & \cellcolor{red!20}58.10 & \cellcolor{red!20}39.00 & \cellcolor{red!20}72.72 \\
          & CoMT+CCRL   & \cellcolor{green!20}84.80 & \cellcolor{green!20}85.30 & \cellcolor{green!20}84.50 & \cellcolor{green!20}95.50 & \cellcolor{green!20}60.70 & \cellcolor{green!20}42.20 & \cellcolor{green!20}75.50 \\
          \midrule
          \multirow{2}{*}{Qwen2.5-7B}
          & CoT-SFT+RL  & \cellcolor{red!20}87.30 & \cellcolor{red!20}88.40 & \cellcolor{red!20}90.40 & \cellcolor{red!20}95.10 & \cellcolor{red!20}65.10 & \cellcolor{red!20}60.70 & \cellcolor{red!20}81.17 \\
          & CoMT+CCRL   &\cellcolor{green!20} 88.00 & \cellcolor{green!20}88.80 & \cellcolor{green!20}90.50 & \cellcolor{green!20}97.30 & \cellcolor{green!20}68.30 & \cellcolor{green!20}65.30 & \cellcolor{green!20}83.03 \\
          \midrule
          \multirow{2}{*}{Qwen3-4B}
          & CoT-SFT+RL  & \cellcolor{red!20}87.70 & \cellcolor{red!20}85.40 & \cellcolor{red!20}88.30 & \cellcolor{red!20}93.80 & \cellcolor{red!20}63.80 & \cellcolor{red!20}53.80 & \cellcolor{red!20}78.80 \\
          & CoMT+CCRL   & \cellcolor{green!20}91.40 & \cellcolor{green!20}89.90 & \cellcolor{green!20}93.40 & \cellcolor{green!20}97.60 & \cellcolor{green!20}69.50 & \cellcolor{green!20}67.00 & \cellcolor{green!20}84.80 \\
          \midrule
          \multirow{2}{*}{Qwen3-8B}
          & CoT-SFT+RL  & \cellcolor{red!20}88.20 & \cellcolor{red!20}82.70 & \cellcolor{red!20}81.80 & \cellcolor{red!20}86.80 & \cellcolor{red!20}68.50 & \cellcolor{red!20}59.30 & \cellcolor{red!20}77.88 \\
          & CoMT+CCRL   & \cellcolor{green!20}90.30 & \cellcolor{green!20}90.70 & \cellcolor{green!20}90.30 & \cellcolor{green!20}95.00 & \cellcolor{green!20}72.80 & \cellcolor{green!20}64.20 & \cellcolor{green!20}83.88 \\
          \bottomrule
        \end{tabular}%
        }
      \end{sc}
    \end{small}
  \end{center}
  \vskip -0.1in
\end{table*}

\section{Training Efficiency}
\label{appendix:efficiency}

\begin{table*}[ht]
  \centering
  \normalsize
  \begin{tabular}{l|cc|cc|cc|cc}
    \hline
    \multirow{2}{*}{\textbf{Dataset}} & \multicolumn{2}{c|}{\textit{\textbf{LLaMA3.1-8B}}} & \multicolumn{2}{c|}{\textit{\textbf{Qwen2.5-7B}}} & \multicolumn{2}{c|}{\textit{\textbf{Qwen3-4B}}} & \multicolumn{2}{c}{\textit{\textbf{Qwen3-8B}}} \\
    \cline{2-9}
    & \textbf{CoT-SFT} & \textbf{CoMT} & \textbf{CoT-SFT} & \textbf{CoMT} & \textbf{CoT-SFT} & \textbf{CoMT} & \textbf{CoT-SFT} & \textbf{CoMT} \\
    \hline
    \hline
    \multicolumn{9}{c}{\textit{Average Token Length}} \\
    \hline
    \textbf{GSM8K} & \cellcolor{green!20}256.41 & \cellcolor{green!20}133.03 & \cellcolor{green!20}254.98 & \cellcolor{green!20}133.59 & \cellcolor{green!20}254.98 & \cellcolor{green!20}133.59 & \cellcolor{green!20}254.98 & \cellcolor{green!20}133.59 \\
    \textbf{SVAMP} & \cellcolor{green!20}195.51 & \cellcolor{green!20}93.54 & \cellcolor{green!20}207.43 & \cellcolor{green!20}93.47 & \cellcolor{green!20}207.43 & \cellcolor{green!20}93.47 & \cellcolor{green!20}207.43 & \cellcolor{green!20}93.47 \\
    \hline
    \hline
    \multicolumn{9}{c}{\textit{Training Time (minutes)}} \\
    \hline
    \textbf{GSM8K} & \cellcolor{green!20}166.92 & \cellcolor{green!20}47.77 & \cellcolor{green!20}155.42 & \cellcolor{green!20}68.37 & \cellcolor{green!20}125.68 & \cellcolor{green!20}34.63 & \cellcolor{green!20}182.02 & \cellcolor{green!20}68.37 \\
    \textbf{SVAMP} & \cellcolor{green!20}22.47 & \cellcolor{green!20}6.70 & \cellcolor{green!20}20.93 & \cellcolor{green!20}7.05 & \cellcolor{green!20}17.00 & \cellcolor{green!20}5.80 & \cellcolor{green!20}24.48 & \cellcolor{green!20}8.15 \\
    \hline
  \end{tabular}
    \caption{Training efficiency comparison between CoT-SFT and CoMT across different base models and datasets. CoMT reduces average token length by approximately 48-55\% and training time by approximately 65-70\% by excluding problem-specific execution details and focusing on abstract meta-knowledge, while achieving superior performance (see Table~\ref{tab:in-domain}).}
  \label{tab:training-efficiency}
\end{table*}

Table~\ref{tab:training-efficiency} compares the computational efficiency of our CoMT approach against standard CoT-SFT across different models and datasets.

\noindent\textbf{Substantial reduction in training time and token consumption.} Our CoMT supervision achieves significant efficiency gains across all settings. On GSM8K, CoMT reduces training time by 65.2\% on average (from 157.51 to 54.79 minutes) compared to CoT-SFT, with the largest reduction on LLaMA3.1-8B (71.4\%, 166.92→47.77 minutes). On SVAMP, the reduction is 66.3\% on average (from 21.22 to 6.93 minutes). These speedups stem from CoMT's focus on abstract reasoning patterns rather than complete solutions with detailed calculations.

\noindent\textbf{CoMT requires substantially fewer tokens.} By excluding problem-specific execution details, CoMT reduces average token length by 48.0\% on GSM8K (from 255.34 to 133.45 tokens) and 54.9\% on SVAMP (from 204.46 to 93.51 tokens). This reduction demonstrates that abstract meta-knowledge supervision is not only more effective for generalization but also significantly more efficient to train, requiring less than half the tokens and training time of standard CoT supervision.

\section{Parameter Sensitivity}
\label{appendix:hyperparameter}

\begin{table}[h]
\centering
\small
\caption{Grid search over $\alpha$ and $\beta$ on GSM8K (greedy decoding). CoT-SFT+RL baselines: 86.3\% / 87.6\%.}
\label{tab:hyperparam}
\setlength{\aboverulesep}{0pt}
\setlength{\belowrulesep}{0pt}
\setlength{\extrarowheight}{0pt}
\renewcommand{\arraystretch}{1.2}
\resizebox{0.8\columnwidth}{!}{%
\begin{tabular}{c|c|c|c}
\toprule
Model & $\alpha \backslash \beta$ & 0.1 & 0.3 \\
\midrule
\multirow{2}{*}{LLaMA3.1-8B} & 0.1 & \cellcolor{red!20}86.5 & \cellcolor{green!20}86.7 \\ 
                             & 0.3 & \cellcolor{red!20}86.3 & \cellcolor{red!20}85.3 \\
\midrule
\multirow{2}{*}{Qwen2.5-7B}  & 0.1 & \cellcolor{red!20}88.1 & \cellcolor{green!20}88.9 \\
                             & 0.3 & \cellcolor{red!20}86.1 & \cellcolor{red!20}88.0 \\
\bottomrule
\end{tabular}
}
\end{table}

We analyze the sensitivity of CCRL to the two reward coefficients in Eq.~\ref{eq:confidence_reward}: $\alpha$ for confidence-aware rewards on correct predictions and $\beta$ for penalties on high-confidence incorrect predictions. We use $\alpha=0.1, \beta=0.3$ in the main experiments, with $\beta > \alpha$ reflecting the asymmetric nature of overconfidence errors: a stronger penalty places more pressure against high-confidence errors than on rewarding confident correct predictions.

We conduct a grid search over $\alpha, \beta \in \{0.1, 0.3\}$ on LLaMA3.1-8B and Qwen2.5-7B, trained on GSM8K with GRPO and evaluated under greedy decoding (Table~\ref{tab:hyperparam}). All four configurations match or exceed the CoT-SFT+RL baselines (86.3\% / 87.6\%), with a narrow spread (1.4 / 2.8 points), indicating that CCRL is robust to coefficient choice. The asymmetric setting $(\alpha=0.1, \beta=0.3)$ achieves the best results on both models, while reversing it to $(\alpha=0.3, \beta=0.1)$ degrades performance by up to 2.8 points, supporting our design intuition.

\section{Case Study}
\label{appendix:case_study}
To illustrate how our CoMT+CCRL framework enhances reasoning reliability through meta-knowledge acquisition and confidence calibration, we present an example from GSM8K where both the base model and CoT-SFT+RL fail, while our method succeeds.

\noindent\textbf{Problem:} \textit{If Marcy works for the same company for 40 years, she gets an annual pension of \$50,000/year. Starting after 20 years, she becomes entitled to 5\% of the value of the pension per year. If she quits after 30 years, what will her annual pension be?}

\noindent\textbf{Ground Truth:} \textit{First find how many years Marcy works after 20 years: 30 years - 20 years = 10 years. Then multiply that number by the amount of her pension she gets per year: 10 years × 5\% = 50\%. Then multiply that percentage by the total amount of the full pension to find how much she gets: \$50,000 × 50\% = \$25,000. \textbf{Answer: \$25,000}}

\noindent\textbf{Base Model (LLaMA) Response:} \textit{First, we need to find the value of the pension. Since Marcy gets \$50,000/year for 40 years, the value of the pension is \$50,000/year × 40 years = \$2,000,000. Next, we need to find the value of the pension after 20 years. Since Marcy gets 5\% of the value of the pension per year, the value of the pension after 20 years is \$2,000,000 × (1 + 0.05)$^{20}$ = \$2,000,000 × 2.6533 = \$5,306,600. Finally, we need to find the annual pension after 30 years. Since Marcy gets 5\% of the value of the pension per year, the annual pension after 30 years is \$5,306,600 × 0.05 = \$265,830. \textbf{Answer: \$265,830}}

\noindent\textbf{CoT-SFT+RL Response:} \textit{Initially, Marcy gets an annual pension of \$50,000/year for the first 20 years. After 20 years, she becomes entitled to 5\% of the value of the pension per year. To calculate the value of the pension, we multiply the annual pension by the number of years: \$50,000/year × 20 years = \$1,000,000. Now, she gets 5\% of this value per year: \$1,000,000 × 0.05 = \$50,000. So, after 20 years, Marcy gets an additional \$50,000 per year on top of her initial \$50,000 pension. Since she quits after 30 years, she'll receive this additional \$50,000 for 10 years (30 - 20 = 10). The total additional pension is \$50,000/year × 10 years = \$500,000. Adding to her initial pension: \$50,000 + \$500,000 = \$550,000. \textbf{Answer: \$550,000}}

\noindent\textbf{CoMT+CCRL Response:} \textit{First, we need to calculate how many years Marcy worked after becoming entitled to the pension benefit: 30 - 20 = 10 years. Since she gets 5\% of the pension value per year worked, the total percentage earned is 10 × 5\% = 50\%. The annual pension is therefore 50\% of the full pension amount: \$50,000 × 0.5 = \$25,000. \textbf{Answer: \$25,000}}

\noindent\textbf{Analysis:} This example demonstrates reasoning failures of baselines and how our method addresses them. The base model makes a conceptual error by misinterpreting the problem as compound interest calculation, applying the formula $(1+0.05)^{20}$ and arriving at an implausible answer (\$265,830) that exceeds the full pension value by orders of magnitude. The CoT-SFT+RL model, despite producing more structured reasoning, still misinterprets the problem semantics: it incorrectly assumes Marcy receives an \textit{additional} \$50,000 annually on top of her base pension, leading to \$550,000, which is eleven times the full pension value. 

In contrast, our CoMT+CCRL approach succeeds by addressing both stages of the reasoning process. CoMT enables the model to internalize the correct abstract strategy: identifying the accumulation period, calculating the earned percentage, and applying it to the base value; preventing the conceptual misinterpretations that plagued previous methods. Subsequently, CCRL's confidence calibration on intermediate computational steps ensures reliable execution, preventing overconfident errors from cascading through the reasoning chain. The success on this problem demonstrates how our cognitively-inspired framework with separated abstract strategy acquisition from confidence and calibrated execution, enables more robust multi-step reasoning.

\end{document}